\documentclass{article}

\usepackage[final]{neurips_2024}

\usepackage{fancyhdr}

\usepackage[utf8]{inputenc} 
\usepackage[T1]{fontenc}    
\usepackage{hyperref}       
\usepackage{url}            
\usepackage{booktabs}       
\usepackage{amsfonts}       
\usepackage{nicefrac}       
\usepackage{microtype}      
\usepackage{xcolor}         
\usepackage{mdframed}
\usepackage{listings}
\usepackage{algorithm}
\usepackage{algorithmic}
\usepackage{graphicx}
\usepackage{multirow}
\usepackage{tikz}
\usepackage{subcaption}
\usetikzlibrary{shapes, arrows.meta, positioning}
\usepackage{amsmath, amssymb, amsthm}
\usepackage{graphicx}
\usepackage{hyperref}
\usepackage{algorithm}
\usepackage{algorithmic}
\usepackage{listings}
\usepackage{tcolorbox}
\usepackage{booktabs}
\newtheorem{theorem}{Theorem}
\usepackage{caption} 
\usepackage{subcaption} 
\usepackage{pgfplots}
\usepackage[verbose]{placeins}
\usepackage{seqsplit}
\usepackage[utf8]{inputenc} 
\usepackage[T1]{fontenc}    
\usepackage{hyperref}       
\usepackage{url}            
\usepackage{booktabs}       
\usepackage{amsfonts}       
\usepackage{nicefrac}       
\usepackage{microtype}      
\usepackage{xcolor}         
\usepackage{tocloft}
\usepackage{titletoc}
\usepackage{appendix}
\fancypagestyle{plain}{
  \fancyfoot[C]{}
  \fancyhead{}

}
\newlength\myindent
\setcitestyle{aysep={},citesep={;}}

\title{Proving Olympiad Algebraic Inequalities without Human Demonstrations}

\author{%
  \textbf{Chenrui Wei}\textsuperscript{1}\\
  \texttt{chenruiw97@gmail.com}
  \And \textbf{Mengzhou Sun}\textsuperscript{2}\\
  \texttt{sunm07@u.nus.edu}
  \And \textbf{Wei Wang}\textsuperscript{1,} \thanks{Corresponding author.}\\
  \texttt{wangwei@bigai.ai}
}

\begin{document}

\maketitle
\vspace{-0.8cm}
\begin{center}
\textsuperscript{1}State Key Laboratory of General Artificial Intelligence, BIGAI, Beijing, China \\
\textsuperscript{2}Department of Mathematics, National University of Singapore
\end{center}
\vspace{0.2cm}
\begin{abstract}
Solving Olympiad-level mathematical problems represents a significant advancement in machine intelligence and automated reasoning. Current machine learning methods, however, struggle to solve Olympiad-level problems beyond Euclidean plane geometry due to a lack of large-scale, high-quality datasets. The challenge is even greater in algebraic systems, which involve infinite reasoning spaces within finite conditions. To address these issues, we propose \textit{AIPS}, an \textit{Algebraic Inequality Proving System} capable of autonomously generating complex inequality theorems and effectively solving Olympiad-level inequality problems without requiring human demonstrations. During proof search in a mixed reasoning manner, a value curriculum learning strategy on generated datasets is implemented to improve proving performance, demonstrating strong mathematical intuitions. 
On a test set of 20 International Mathematical Olympiad-level inequality problems, AIPS successfully solved 10, outperforming state-of-the-art methods. Furthermore, AIPS automatically generated a vast array of non-trivial theorems without human intervention, some of which have been evaluated by professional contestants and deemed to reach the level of the International Mathematical Olympiad. Notably, one theorem was selected as a competition problem in a major city's 2024 Mathematical Olympiad.
All the materials are available at {\it \href{https://sites.google.com/view/aips2}{sites.google.com/view/aips2}}.

\end{abstract}

\section{Introduction}

One of the key milestones in the field of artificial intelligence is the capability to reason \citep{pearl1998graphical} and prove theorems \citep{wu1978decision, chou2000deductive, trinh2024}. However, theorem proving often involves long reasoning chains, complex mathematical structures, intricate calculations, and infinite reasoning spaces. Consequently, developing AI capable of proving complex mathematical theorems requires sophisticated reasoning and the ability to navigate through an extensive search space to construct a valid proof. 
The complexity of these problems lies in the need for effective heuristics and strategies to manage the vast number of possible actions and the lengthy sequences of logical steps necessary to arrive at a solution.

Existing work on grade school and college admission math problems has achieved notable success, e.g., GSM8K \citep{gsm8k} and SAT Math \citep{gpt4tr}, which demonstrate better performance on tasks such as arithmetic and basic algebra. However, research focused on solving International Mathematical Olympiad (IMO)-level problems remains relatively sparse. Notable efforts in this area include AlphaGeometry \citep{trinh2024}, and GPT-{\it{f}} \citep{polu2020generative} on miniF2F \citep{minif2f}, which have made progress in solving Euclidean plane geometry at the Olympiad level and various mathematical competition problems, respectively.

A significant challenge for learning-based methods in this domain is the scarcity of suitable datasets, which limits the ability to train models effectively and hampers progress in achieving human-level performance on these high-difficulty problems. 
The miniF2F dataset \citep{minif2f} includes only 244 validation and 244 test
mathematical problems from various competitions. AlphaGeometry \citep{trinh2024} addresses this issue by synthesizing millions of theorems and proofs across different levels of complexity to train a neural language model from scratch.
Similarly, the INequality Theorem proving benchmark, INT \citep{wu2020int}, can synthesize a theoretically unlimited number of theorems and proofs in the domain of algebraic equalities and inequalities. However, INT focuses on testing a learning-assisted theorem proving agent's generalization ability rather than increasing the difficulty to competition level.

Another significant challenge in automated theorem proving is designing effective search strategies to navigate the vast space of possible proofs. Recent advancements have highlighted various approaches to enhance search efficiency and proof success rates. Some studies have shown that incorporating Monte Carlo Tree Search (MCTS) at test time can significantly aid in proving new theorems \citep{wu2020int}. Inspired by the success of AlphaZero \citep{zhang2020alphazero}, other research has explored HyperTree Proof Search (HTPS) \citep{lample2205hypertree}, which learns from previous proof searches through online training, iteratively improving its strategy by learning which paths are more likely to lead to successful proofs. Another innovative approach starts the proof search from the root goal that needs to be proved \citep{polu2020generative}, expanding a maintained proof tree by prioritizing open goals based on their cumulative log probability.

In this work, we introduce \textit{AIPS}, an \textit{Algebraic Inequality Proving System}, which can generate a large number of high-quality theorems and solve IMO-level algebraic problems. AIPS focuses on ternary and quaternary inequalities, excluding $n$-variable inequalities represented recursively in formal verification systems. Among the generated theorems, some have proven to be very challenging, with one selected for a major city’s 2024 Mathematical Olympiad. We present novel and challenging inequality theorems discovered by AIPS in the appendix, which have been carefully evaluated by IMO-level professional contestants and found to be comparable to IMO inequalities from around the year 2000.

Additionally, AIPS incorporates a value network to evaluate newly generated inequalities, selecting subgoal candidates based on the top scores provided by the value network. The value network is trained on synthetic datasets with increasing difficulty in a curriculum manner. In our experiments, AIPS proved difficult theorems up to the IMO level and solve 10 out of 20 problems in an IMO-level inequality test, significantly surpassing the performance of previous Large Language Model-based theorem provers \citep{polu2020generative, polu2022formal, leandojo, leancopilot}.

The main contributions in this paper are summarized as follows:
\begin{enumerate}
\item We propose a symbolic deductive engine capable of efficiently generating high-quality and solving high-difficulty algebraic inequality theorems. This engine addresses the bottleneck of lacking large-scale, high-quality data in this field. 
\item We demonstrate that a symbolic algebraic inequality prover can be significantly enhanced under the guidance of a value network, especially when the value network is trained in a curriculum manner.
\item Our AIPS can generate challenging and elegant inequality theorems, including one selected for a major city’s Mathematical Olympiad. AIPS proves 10 out of 20 IMO-level inequalities, surpassing state-of-the-art methods and producing highly human-readable proofs.
\end{enumerate}

\section{Related Work}
\label{gen_inst}

{\bf Automated Theorem Proving.}
Automated theorem proving has been a focus of artificial intelligence since the 1950s \citep{1950atp, wu1978decision}. Modern theorem provers, based on tactic and premise selection, search for proofs by interacting with proof assistants such as Lean \citep{de2015lean}, Coq \citep{barras1999coq} and Isabelle \citep{nipkow2002isabelle}. They struggle with the rapidly expanding search space and the scarcity of high-quality datasets in most mathematical domains. The challenge is even greater for proving algebraic inequalities, which involve complex computational rules. Previous efforts to address this issue have focused on augmenting tactic selection and premise prediction in interactive theorem provers \citep{polu2020generative, polu2022formal, leandojo}. However, these provers have only been able to solve problems of limited difficulty in this field.
In this paper, our AIPS can solve highly complex algebraic inequality theorems up to the IMO level.

{\bf Datasets and Benchmarks for Theorem Proving.}
Formal mathematical libraries, such as Isarstep \citep{li2020isarstep}, Mathlib \citep{mathlib}, and CoqGym \citep{coqgym}, currently serve as the primary datasets for theorem proving. These libraries, manually curated by humans, include many intricate and profound proofs, such as the formal proofs of the Four-Color Theorem \citep{4color}, the Liquid Tensor Experiment \citep{scholze2022liquid}, and Fermat's Last Theorem \citep{fmlast}. Due to the labor-intensive nature of manual proof writing, these libraries are relatively small, typically containing around 200,000 theorems. While they encompass a wide range of mathematical fields, the number of theorems in specific areas is quite limited.

Synthetic theorems can provide large-scale datasets for learning-based theorem provers \citep{polu2020generative, wu2020int}. However, these theorems are often of limited difficulty. Recently, significant progress has been made in synthesizing geometry theorems \citep{trinh2024} using neural theorem provers. In this paper, we develop AIPS for algebraic inequalities, which can automatically and efficiently generate a large number of intricate theorems, with some reaching the IMO level. These theorems will significantly improve neural theorem proving methods.

{\bf Search Strategy for Efficient Inference.}
Deep learning has achieved remarkable success in enhancing search algorithms \citep{silver2016mastering, silver2017mastering}. Proof search in theorem proving, however, is more challenging compared to self-play games like Go, as it may involve an infinite search space within finite conditions. INT \citep{wu2020int} incorporates MCTS, while HyperTree Proof Search (HTPS) \citep{lample2205hypertree} employs online training to improve search strategy. GPT-{\it{f}} \citep{polu2020generative} learns a value network to guide backward search. Our AIPS integrates the benefits of both HTPS and GPT-{\it{f}}, introducing a value curriculum learning strategy.

\section{Algebraic Inequality Proving System}


\subsection{Symbolic Deductive Engine for Algebra}
\label{headings}

Interactive theorem provers, such as Lean, can verify mathematical operations but lack the ability to perform automated mathematical reasoning by combining computational rules. This challenge is amplified in the automatic proof of algebraic inequalities, which often involves numerous calculations, extensive transformation rules, and complex theorem matching. To address this, we design a symbolic deductive engine for algebra, encompassing dozens of fundamental theorems and transformation rules for algebraic inequalities. It integrates with the symbolic computation system SymPy \footnote{https://www.sympy.org/}, enabling effective algebraic reasoning.




\subsubsection{Representation for Algebraic Expressions and Theorems}

\begin{figure}[h!]
    \centering
    \includegraphics[width=1.0\textwidth]{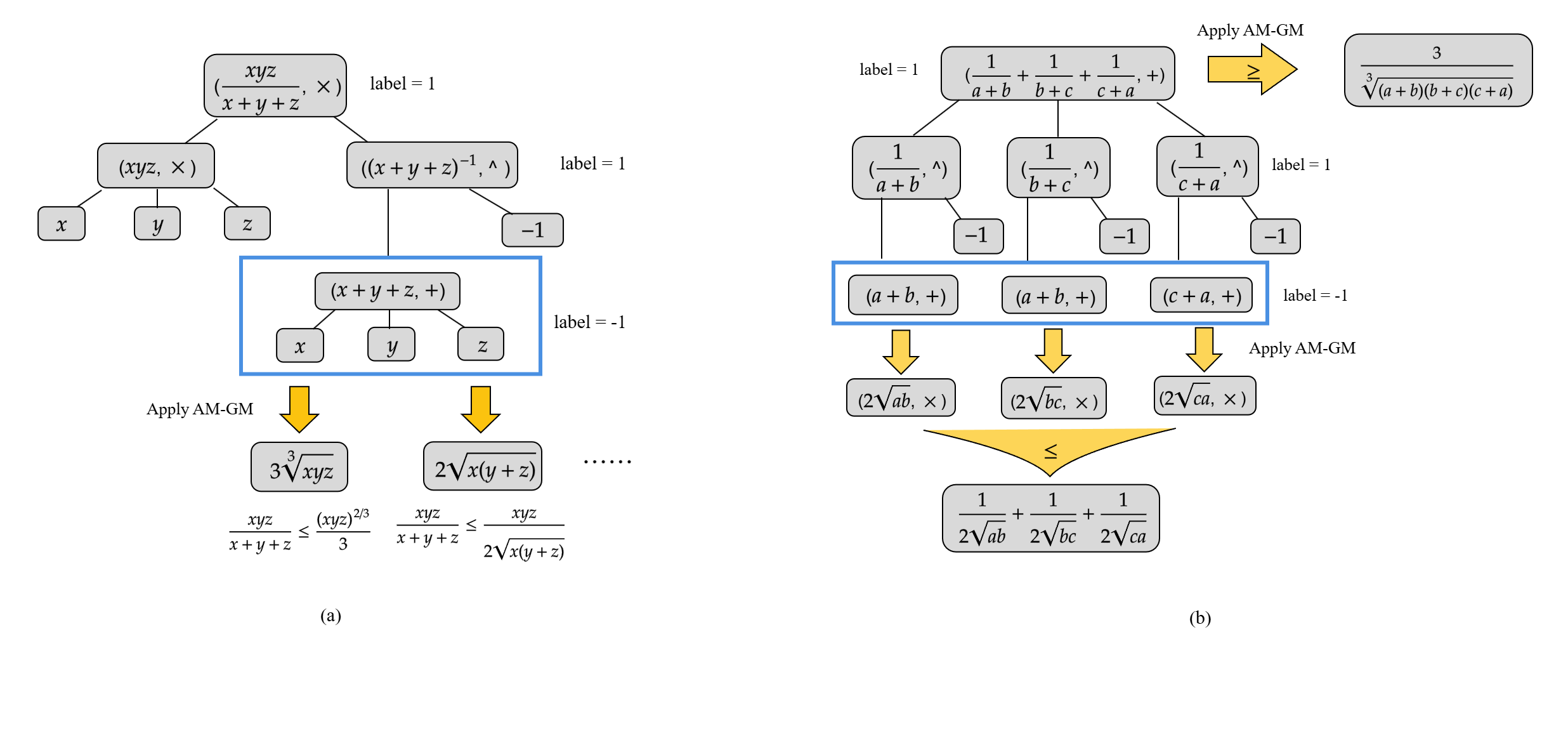}
    \caption{Examples of expression trees and pattern matching for the AM-GM inequality are illustrated. In (a), for $x, y, z \geq 0$, the value of $\frac{xyz}{x+y+z}$ decreases as $x+y+z$ increases, so the label of the node $x+y+z$ is $-1$. By applying the AM-GM inequality, we derive a series of upper bounds with respect to the root, e.g., $\frac{(xyz)^{2/3}}{3}$ and $\frac{xyz}{2\sqrt{x(y+z)}}$. In (b), when traversing the expression tree of $\frac{1}{a+b} + \frac{1}{b+c} + \frac{1}{c+a}$, pattern matching for the AM-GM inequality at various nodes yields different types of bounds, such as the upper bound $\frac{1}{2\sqrt{ab}} + \frac{1}{2\sqrt{bc}} + \frac{1}{2\sqrt{ca}}$ and the lower bound $\frac{3}{((a+b)(b+c)(c+a))^\frac{1}{3}}$.}
    \label{fig:exprtree}
\end{figure}

Algebraic expressions are represented symbolically with an underlying expression tree structure as shown in Fig. \ref{fig:exprtree}. The basic computational rules include self-equivalence transformations of inequalities and various built-in SymPy functions, such as combining fractions (\texttt{sympy.together}) and expanding expressions (\texttt{sympy.expand}). Our deductive engine's library also includes fundamental algebraic inequality theorems: the Arithmetic Mean-Geometric Mean Inequality (AM-GM), the weighted AM-GM Inequality, Cauchy's Inequality, Jensen's Inequality, the discrete Hölder's Inequality, Schur's Inequality, the binary and ternary Muirhead's Theorem. Each inequality is represented as a category of theorem matching, containing variables, conditions, conclusions, and equality conditions.

\subsubsection{Pattern Matching for Inequality Theorems}

During symbolic reasoning, the system attempts to apply inequality theorems to a particular algebraic expression or inequality, as shown in Fig. \ref{fig:exprtree}. When matching algebraic expressions with inequality theorems, it first traverses the expression tree to determine how the value of the entire expression changes as the node's value increases, updating the node's label accordingly. If the change cannot be determined, no theorem matching is performed on the subtree of that node. After completing the labeling, the system matches the next layer of determinable nodes with theorems. If a match is successful, the matched sub-expression is replaced with the new expression obtained using the theorem. Based on the previous labels, it then determines whether the entire expression increases or decreases, thereby deriving a new inequality. For certain inequality theorems, such as Jensen's Inequality, pattern matching is particularly complex and time-consuming. Therefore, to improve the efficiency of reasoning at each step, we have imposed time limits on the matching process for some theorems.

\subsubsection{Forward Reasoning}
\label{fwd_reasoning}

Forward reasoning in theorem proving involves matching variables and conditions to a theorem and deducing new conclusions. In our engine, new inequalities can be obtained by matching theorems to both sides of an inequality or by applying self-equivalence transformation rules. If any two of the resulting inequalities can be connected (e.g., applying \(a \leq b\) and \(b \leq c\) to derive \(a \leq c\)), the system continues to link them to form new inequalities. Therefore, our engine has the capability to perform forward reasoning to generate large-scale data.

\subsection{Olympiad-Level Inequality Proof Set}
\label{chap_4}

One of the main challenges in enabling learning-based models to solve complex mathematical problems is the scarcity of large-scale, high-quality datasets. To overcome this obstacle, we develop a theorem generator that effectively generates Olympiad-level inequality theorems by enhancing the methods described in Section \ref{fwd_reasoning}.

\begin{figure}[h!]
    \centering
    \includegraphics[width=0.8\textwidth]{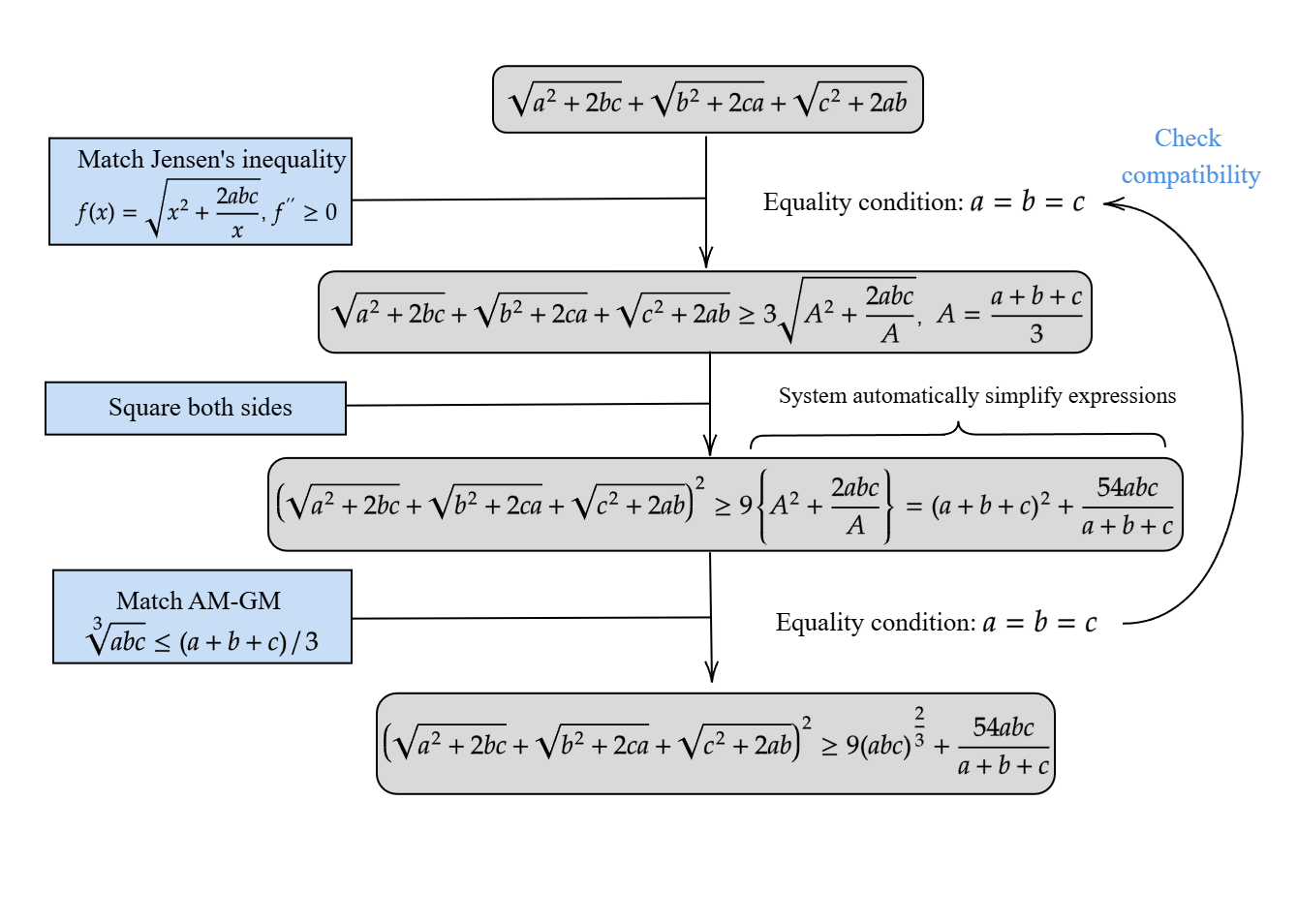}
    \caption{An example of generating synthetic theorems in AIPS. When the initial premise $\sqrt{a^2+2bc}+\sqrt{b^2+2ca}+\sqrt{c^2+2ab}$ successfully matches with Jensen's inequality, a new inequality is generated. By subsequently applying transformation rules and matching other fundamental inequalities, such as the AM-GM inequality, the deductive engine incrementally generates new inequality theorems. When an inequality theorem is applied, the system verifies whether the equality condition holds, e.g., $a=b=c$.}
    \label{fig:data-gen}
\end{figure}

\subsubsection{Synthetic Theorem Generation}
We randomly generate thousands of cyclically symmetric symbolic expressions, which serve as the initial premises for our reasoning process. Utilizing 32 CPUs, we run Algorithm \ref{alg:generate_theorems} for 8 hours, resulting in the generation of 191,643 inequality theorems. The generated inequalities are stored in a tree structure, with each node containing the necessary information for extracting proofs and training machine learning models. Fig. \ref{fig:data-gen} shows the procedure of generating a synthetic theorem in our AIPS, and Fig. \ref{fig:rome}(a) shows the distribution of inference depths in the generated inequalities. 

\begin{algorithm}
\caption{Generating Theorems}
\label{alg:generate_theorems}
\begin{algorithmic}[1]
\STATE \textbf{function} \texttt{Generate\_Theorems}(\textit{expression} $P$, \textit{loops} $N$)
    \begin{ALC@g}
        \STATE Initialize \textit{Theorem Set} $S$, 
    
        \textit{Inequality Transformation Rules} $O$, \textit{Inequality Sets} $A1$, $A2$, $A3$
        \STATE Apply $S$ to $P$ to obtain a series of inequalities and add those whose equality conditions hold to a set $R$
        \FOR{$i \leftarrow 1$ to $N$}
            \FOR{each inequality $ineq$ in $R$}
                \STATE Apply rules $O$ to $ineq$ to obtain $A1$
            \ENDFOR
            \FOR{each inequality $ineq$ in $R$}
                \STATE Apply theorems $S$ to one side of $ineq$ and check if it can be linked to the original inequality. If so, add it to $A2$
            \ENDFOR
            \FOR{each inequality $ineq$ in $A2$}
                \STATE Check if $ineq$ meets the equality condition and add it to $A3$ if it does
            \ENDFOR
            \STATE Update $R$ by selecting $M$ inequalities from the union of $A3$ and $A1$ according to the length of inequalities
        \ENDFOR
        \STATE \textbf{return} $R$
    \end{ALC@g}
\STATE \textbf{end function}
\end{algorithmic}
\end{algorithm}

\subsubsection{Synthetic Theorem Evaluation}

To evaluate the quality of our dataset, we select 10 problems with reasoning lengths exceeding five steps, and invite two National Mathematical Olympiad gold medalists and one silver medalist to assess the difficulty and elegance of these problems. Their evaluations reveal that our dataset contains a vast array of non-trivial theorems, some of which surpass the difficulty of inequalities found in early IMO competitions. Notably, one inequality theorem from our dataset is selected for a major city's Mathematical Olympiad. All the 10 problems and evaluation results are provided in Appendix \ref{syn-thms}.

\subsection{Neural Algebraic Inequality Prover}

\begin{figure}[t!]
\centering
\begin{subfigure}{0.40\textwidth}
    \includegraphics[width=\textwidth]{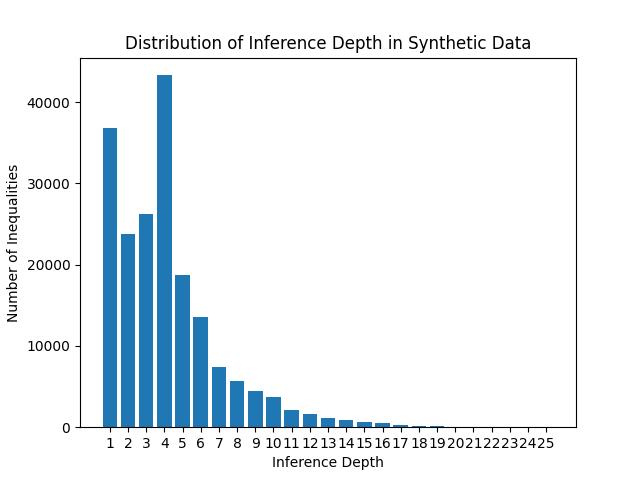}
    \caption{Distribution of inference depths. In the process of generating synthetic theorems, we limit the reasoning steps. Unlike geometry problem, long reasoning chains in inequality generation can lead to trivial theorems. Solutions to challenging IMO inequalities typically involve only two or three steps of matching inequality theorems.}
    \label{fig:first}
\end{subfigure}
\hfill
\begin{subfigure}{0.56\textwidth}
    \includegraphics[width=\textwidth]{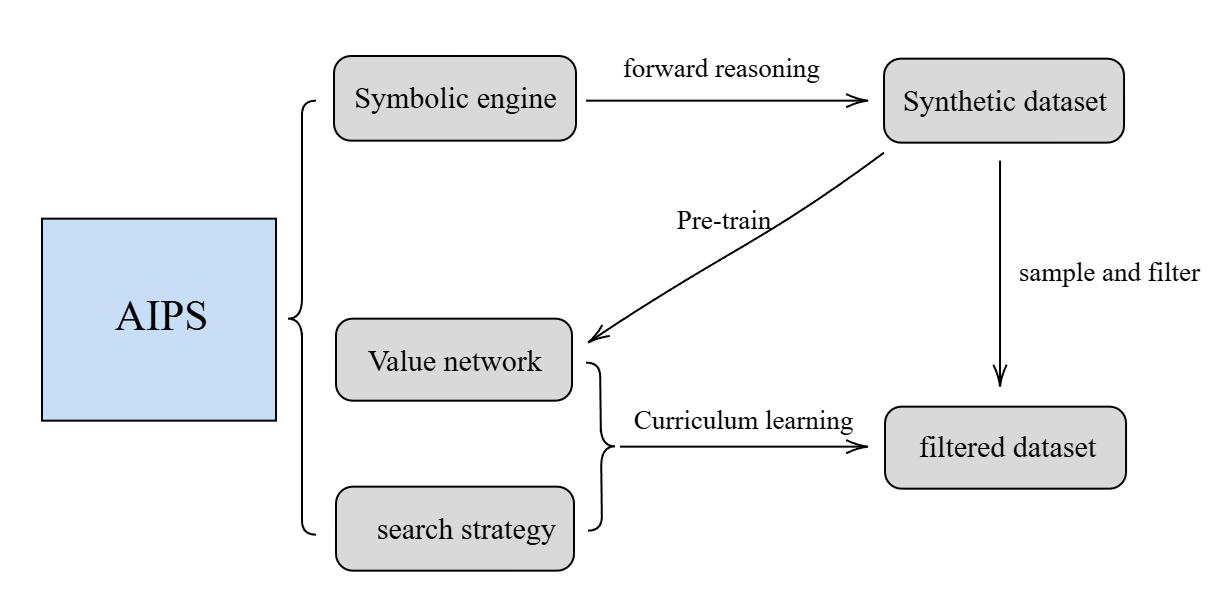}
    \caption{Self-evolving process of AIPS. After pre-training on the initial synthetic dataset, AIPS is capable of proving some challenging theorems. Guided by the value network, it then attempts to solve problems in an increasingly difficult filtered dataset. By extracting nodes on the proof path as positive labels and other nodes as negative labels, it fine-tunes the value network and gradually improves proving performance in a curriculum manner.}
    \label{fig:second}
\end{subfigure}
\vspace{-5pt}
\caption{(a) Distribution of inference depths in our dataset. (b) Self-evolving process of AIPS.}
\label{fig:rome}
\vspace{-10pt}
\end{figure}

\begin{figure}[h!]
    \centering
    \includegraphics[width=0.95\textwidth]{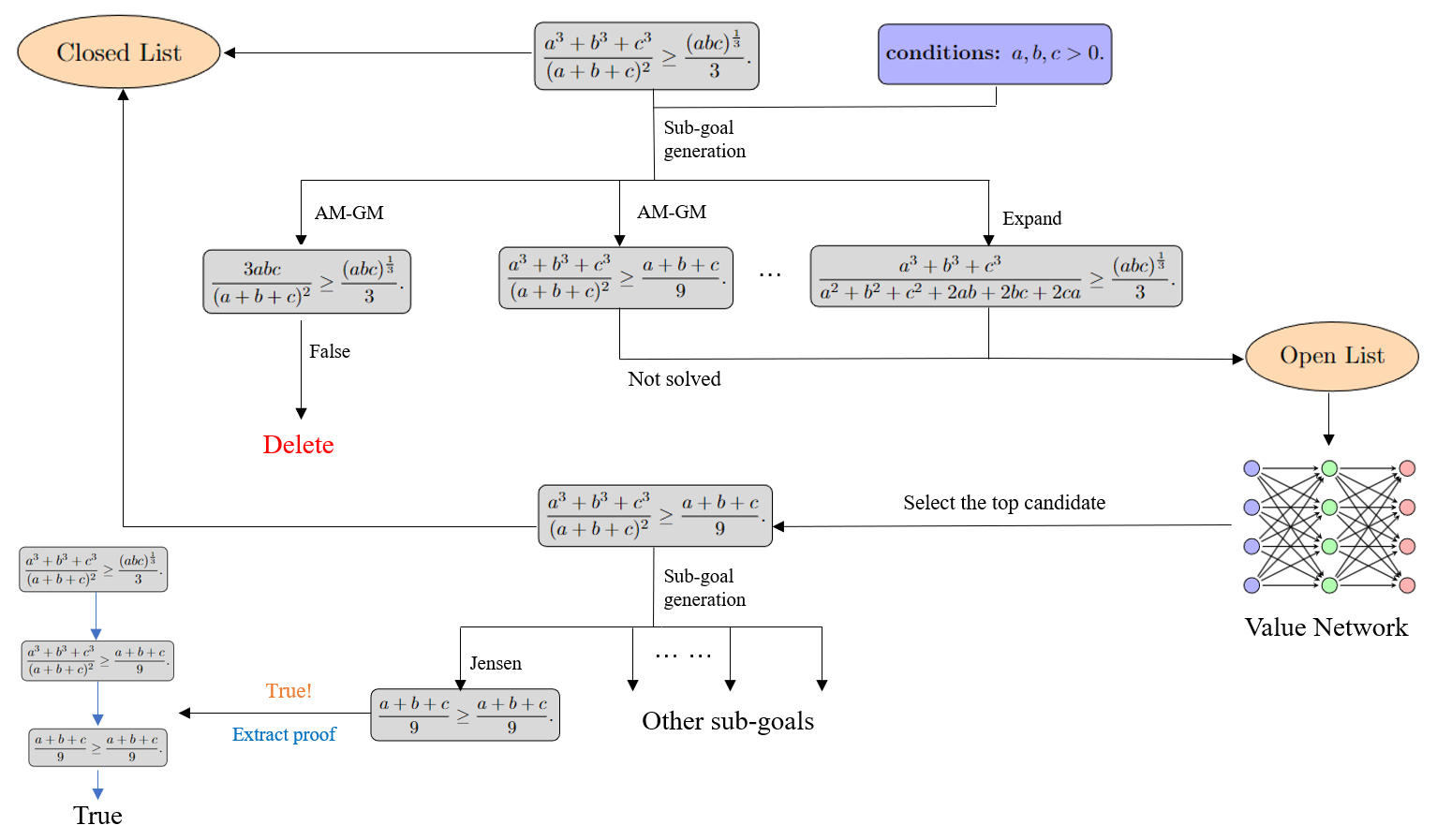}
    \caption{Overview of how AIPS proves a simple theorem. At each step, the deductive engine attempts to match inequality theorems with each side of the goal and applies all transformation rules to the expression, resulting in a list of new subgoals. The searched goal is placed into a closed list, ensuring that it will not be examined again. If one of the new subgoals is true, indicating that the inequality holds, then the theorem is proved. Otherwise, the new subgoals are added to the open list, along with other subgoals generated previously. A value network then evaluates all subgoals in the open list, and the top-value one is chosen for the next iteration of proof search.}
    \label{fig:pftree}
\end{figure}

By leveraging the capabilities of the deductive engine introduced in Section \ref{headings} and the Best-First-search algorithm \citep{bestfs}, we develop an algebraic inequality prover. This prover formulates the algebraic inequality proving as a sequential decision-making process by selecting theorems to generate highly human-readable proofs. As shown in Fig. \ref{fig:pftree}, given a goal and related conditions, AIPS first generates a list of subgoals by applying a set of theorems at each iteration. A value neural network is then used to evaluate these newly generated subgoals along with the previous unresolved subgoals. The top-value subgoal is selected for the next step of reasoning. This iterative process continues until the proof is successfully completed, as shown in Fig. \ref{fig:rome}(b).

\subsubsection{Searching Proofs by Combining Value Network with Symbolic Prover}
\label{search-pf}
The procedure of searching for inequality proofs is generally divided into three parts: mixed reasoning for subgoal generation, evaluation, and planning.

\textbf{Subgoal Generation.} There are two methods for generating subgoals in AIPS. The first method involves applying fundamental inequality theorems. Let $X$ be the set of variables. Suppose the inequality theorem to prove is \(u(X) \leq v(X)\) under a condition set \(\mathcal{P}\). AIPS first homogenizes the inequality to \(f(X) \leq g(X)\) on both sides by applying conditions in \(\mathcal{P}\). Then, by applying theorems to the left-hand side of the target inequality, AIPS generates a series of new inequalities:
\[
f(X) \leq h_1(X), \ldots, f(X) \leq h_n(X)
\]
This results in subgoals \(h_i(X) \leq g(X)\). Similarly, by applying theorems to the right-hand side, AIPS also generates subgoals \(f(X) \leq s_j(X)\). The second method involves applying transformation rules such as \texttt{sympy.expand} and \texttt{sympy.apart} to the goal, generating subgoals that are equivalent to the original inequality.

\textbf{Evaluation.} AIPS employs a value function \(V_{\theta}\) to assess the difficulty of each inequality. Formally, we have a function \(f\) parameterized by \(\eta\) that encodes the inequality expression \(s\). The encoded embedding vector \(f_{\eta}(s)\) is then fed into a deep neural network \(g_{\phi}\), which outputs a value in the interval [0,1]. We choose \(f\) to be a transformer encoder with average pooling \citep{vaswani2017attention}.

\textbf{Planning.} With the evaluation function \(V_{\theta}\), we use the Best-First search algorithm for planning. We also test the performance of MCTS algorithm, where the result is less satisfactory. 

There are two primary reasons for this. First, the action space for each state is extremely large, leading to explosive growth of the MCTS searching tree. Second, the high cost of reasoning steps makes the simulation step in MCTS nearly impractical, often exceeding time limits.

We also note that our prover can be combined with any heuristic function,  and thus design various baselines in our experiments.

\subsubsection{Pre-training Value Network Using a Heuristic Function}
\label{heurists}

We define the tree-depth \(\mathcal{D}\) of an inequality as the maximum depth of the expression trees on both sides. Proving an algebraic inequality is equivalent to reducing the tree-depth of the inequality to one. We use \(\mathcal{D}\) as the supervision information to train initial heuristic function \(f_{\text{init}}\) in the Best-First search algorithm. That is to say, we pre-train a value network \(V_{\theta}\) as \(f_{\text{init}}\) on the synthetic dataset by utilizing the tree-depth \(\mathcal{D}\).

\subsubsection{Fine-tuning Value Network on Filtered Synthetic Data}
We create a new dataset by removing all inequalities with inference depth less than 4. We then randomly sample 1,200 problems and sort them by tree-depth in ascending order. For inequalities with the same tree-depth, they are sorted by the length of their string representation, with shorter lengths placed first.

The fine-tuning procedure involves sequentially proving these inequalities and updating the parameters of the value network. If an inequality is successfully proved, we record the set of subgoals on the proof path as \(T\) and the set of subgoals that are searched but not on the proof path as \(F\). The values of the elements in \(T\) are scaled down by a factor of \(\epsilon\), while the values of the elements in \(F\) are increased. Using these labels, we perform a training round on the value network \(V_{\theta}\), and then proceed to the next problem. This iterative process is used to adjust the network parameters. See Appendix \ref{v-learning} for more details.

\section{Experiments}
We evaluate AIPS on an Olympiad-level algebraic inequality problem test set. It outperforms the state-of-the-art methods in terms of the number of solved problems, demonstrating the strong algebraic intuitions developed by the learned value network.

\subsection{An Olympiad-Level Inequality Benchmark}
\label{benchmark}
Current benchmarks for Olympiad-level math problems, such as miniF2F \citep{minif2f} and Fimo \citep{fimo}, cover a wide array of topics but often lack a dedicated section for algebraic inequalities. In inequality benchmarks like INT \citep{wu2020int}, the problems are typically of limited difficulty. To address this gap, we collect all ternary and quaternary algebraic inequality problems from IMO since 1990. Additionally, we include challenging problems from IMO shortlists and various national mathematical Olympiads, such as the USAMO, the USA National Team Selection Tests, and the Polish, Japanese, and Korean Mathematical Olympiads, all of which are of comparable difficulty to the IMO. In total, we compile 20 problems for our test set, naming it MO-INT-20 (Math-Olympiad-INequality-Test-20). All problems are checked to ensure they are not in AIPS's training datasets. We also translate the test problems into Lean for subsequent experiments.

\subsection{Comparison Methods}
\label{baselines}
Current theorem provers include interactive theorem provers, large language models capable of generating natural language proofs, and neural symbolic theorem provers. We compare LeanCopilot \citep{leancopilot}, the open-source state-of-the-art interactive theorem prover in Lean. Additionally, we evaluate general large language models like GPT-4, GPT-4 Turbo and Gemini 1.5 Pro, as well as the math-specific language model Llemma-7b \citep{llemma}. For neural symbolic theorem provers, we design various baselines, including our deductive engine paired with breadth-first search and MCTS, our deductive engine equipped with tree-depth in Section \ref{heurists} or LLM heuristics as the value function, and our AIPS with only pretrained value network.

It should be noted that we cannot compare with several existing interactive theorem provers \citep{polu2020generative, polu2022formal}  since these provers are not open source to be reproduced. However, it is reported that these provers can only prove a few early Olympiad inequalities, as detailed in the appendix of their respective papers.

\subsection{Comparison Results and Analysis}
We test 11 different provers on the inequalities in MO-INT-20, with each problem limited to 90 minutes of solving time, consistent with the standard problem-solving time in the IMO. All neural-symbolic provers are tested on a single CPU core (equivalent to 1.5 CPU hours per problem). The comparison results are shown in Table \ref{tab:model_performance}. It can be seen that our AIPS achieves the best performance and solves 10 out of 20 problems.

\begin{table}[htbp]
    \centering
    \caption{Model Performances on the MO-INT-20. {\bf DE denotes our deductive engine}. BFS and MCTS are Breadth-First Search and Monte Carlo Tree Search, respectively.}
    \label{tab:model_performance}
    \begin{tabular}{|l|c|c|}
        \hline
        \multicolumn{1}{|c|}{\textbf{Model Category}} & \textbf{Model} & \textbf{Problems Solved (20)} \\
        \hline
        \multirow{3}{*}{\text{Large Language Models}} & \multicolumn{1}{c|}{Gemini 1.5 Pro} & \multicolumn{1}{c|}{1} \\
        \cline{2-3}
                              & \multicolumn{1}{c|}{GPT-4} & \multicolumn{1}{c|}{0} \\
        \cline{2-3}
                              & \multicolumn{1}{c|}{GPT-4 Turbo} & \multicolumn{1}{c|}{0} \\
        \cline{2-3}
                              & \multicolumn{1}{c|}{Llemma-7b} & \multicolumn{1}{c|}{0} \\
        \hline
        \multicolumn{1}{|c|}{\text{Interactive Theorem Provers}} & \multicolumn{1}{c|}{LeanCopilot (LeanDojo)} & \multicolumn{1}{c|}{0} \\
        \hline
        \multirow{3}{*}{\centering Neural-Symbolic Provers} & \multicolumn{1}{c|}{DE + GPT-4 Turbo's heuristics} & \multicolumn{1}{c|}{6} \\
        \cline{2-3}
                                                 & \multicolumn{1}{c|}{DE + BFS} & \multicolumn{1}{c|}{4} \\
        \cline{2-3}
                                                 & \multicolumn{1}{c|}{DE + MCTS} & \multicolumn{1}{c|}{5} \\
        \cline{2-3}
                                                 & \multicolumn{1}{c|}{DE + tree-depth heuristic function} & \multicolumn{1}{c|}{7} \\
        \cline{2-3}
                                                 & \multicolumn{1}{c|}{AIPS with pretrained value network} & \multicolumn{1}{c|}{7} \\
        \cline{2-3}
                                                 & \multicolumn{1}{c|}{AIPS} & \multicolumn{1}{c|}{10} \\
        \hline
    \end{tabular}
\end{table}

{\bf Analysis of Large Language Models' Performance.}
Large language models like GPT-4 have demonstrated remarkable reasoning abilities \citep{llm-reasoning1, llm-cot}. However, in this test, only one of the four models, Gemini 1.5 Pro, successfully generates a fully correct natural language proof. When solving problems, large language models tend to either make trivial mistakes or indicate that they do not know how to solve them, despite the potential contamination of their training data by online proofs. These results reveal their limited math reasoning ability.


{\bf Analysis on a Formal Theorem Prover's Performance.}
Recent studies reveal the capabilities of neural theorem provers based on Interactive Theorem Prover (ITP) frameworks \citep{leandojo, graph2tac}. These systems generally convert theorem proving into code completion tasks. We evaluate the performance of one such theorem prover, LeanCopilot \citep{leancopilot}, developed from LeanDojo, on our test set. LeanCopilot is the current open-source state-of-the-art theorem prover based on Lean. The results indicate its limited ability to solve complex algebraic problems: None of the problems are solved through proof search in LeanCopilot. Additional tests on tactic suggestions (see Appendix \ref{test-res-lean}) show that current formal theorem provers struggle to predict the complex premises required for proving inequalities.

{\bf Analysis on Neural Symbolic Provers' Performance.}
In this test, neural symbolic provers demonstrate a strong ability to prove algebraic inequalities using best-first search algorithm. By applying either breadth-first search or MCTS algorithm, our deductive engine successfully solves four and five problems, respectively. We also test performance under the guidance of a tree-depth heuristic function and a pre-trained value network using the best-first search algorithm, both of which solve seven problems. Additionally, we prompt GPT-4 Turbo and find it exhibit some algebraic intuition, successfully guiding the deductive engine to solve six problems—two more than the breadth-first search. However, it is worth noting that large language models (LLMs) may occasionally prioritize lengthy and meaningless subgoals. Due to the exponential growth of the number of new inequalities as the width and height of the expression trees increase, it can result in expression strings longer than the LLMs' input context length. For example in problem 4 from the 2014 Japan Mathematical Olympiad, it chooses a very long subgoal at iteration 2, resulting in subgoals at the next iteration being three times longer than its input context length. 

Finally, following a curriculum learning strategy on 1,000 inequality problems, AIPS achieves the best performance, solving 10 out of 20 problems. Among the 10 problems from the IMO or IMO shortlist, it successfully solves five, reaching the average level of IMO contestants. We also test the performances of AIPS after 200, 400, 600, and 800 loops of fine-tuning value network (see Appendix \ref{perform-curr}). The results demonstrate that our value curriculum learning strategy is very effective, with the number of proof search steps significantly decreasing during the training process, and the number of solved problems increasing to 10 ultimately. 

\section{Conclusion}
In conclusion, solving Olympiad-level mathematical problems is a significant milestone in machine intelligence and automated reasoning. The lack of large-scale, high-quality datasets presents a challenge, particularly in algebraic systems. To address this, we propose \textit{AIPS}, an \textit{Algebraic Inequality Proving System}, which autonomously generates complex inequality theorems and effectively solves Olympiad-level inequality problems without human input. Utilizing a value curriculum learning strategy, AIPS demonstrated strong mathematical intuition by solving 10 out of 20 International Mathematical Olympiad-level problems. One of these theorems was selected for a major city’s 2024 Mathematical Olympiad.

In the future, by incorporating more fundamental theorems and operational rules, our AIPS could solve even more complex problems, discover a greater number of non-trivial theorems, and assist mathematicians in solving modern mathematical challenges. However, it currently lacks the ability to autonomously propose and comprehend new definitions. Instead, it relies on handwritten theorems and matching rules, which is time-consuming. Addressing this limitation is a crucial area for future research.

\clearpage
\section{Acknowledgements}

We extend our heartfelt gratitude to the three distinguished contestants—two National Mathematical Olympiad gold medalists and one silver medalist—for their invaluable evaluations of our synthetic theorems. We also express our sincere thanks to their coach Zhibin Liang, whose efforts made this collaboration possible. Furthermore, we deeply appreciate the insightful discussions from Jiajun Song, Yuxuan Wang, and Dr. Chi Zhang at Beijing Institute for General Artificial Intelligence. This work was supported in part by the National Natural Science Foundation of China under Grants 61976214.

\bibliographystyle{unsrtnat} 
\bibliography{References} 

\clearpage
\appendix
\section*{Appendix}
\addcontentsline{toc}{section}{Appendix}
\renewcommand{\contentsname}{Appendix}
\startcontents
\printcontents{ }{1}{\setcounter{tocdepth}{2}}

\section{Technical Details of the Deductive Engine}
We provide more information on AIPS' deductive engine and the training process for the value network. To highlight the reasoning ability and maintain readability of proofs, we avoid using brute-force methods such as augmentation-substitution and Wu's method \citet{wu1978decision}.

\subsection{Background}

\subsubsection{Basic Knowledge in Theorem Proving}
\begin{figure}[h]
    
    \centering
    \begin{minipage}{0.45\textwidth}
        \centering
        \includegraphics[width=\textwidth]{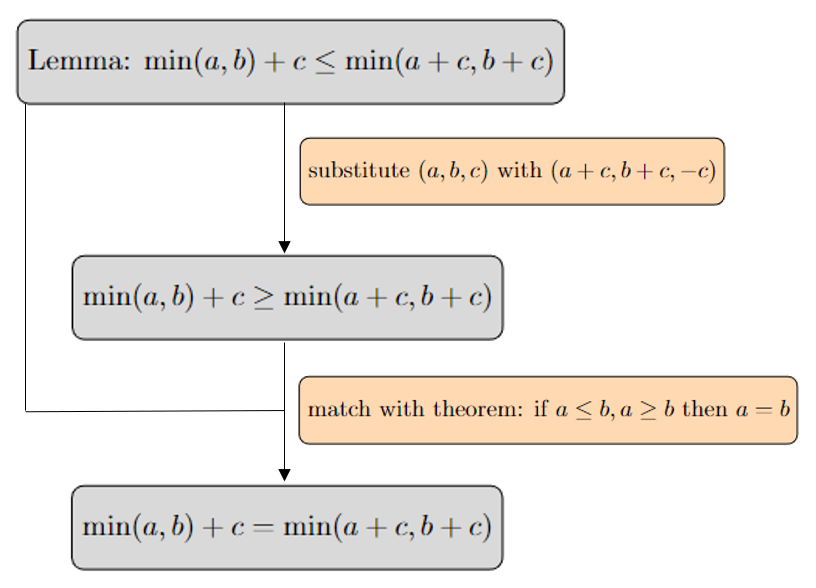}
    \end{minipage}
    \hfill
    \begin{minipage}{0.45\textwidth}
        \centering
        \includegraphics[width=\textwidth]{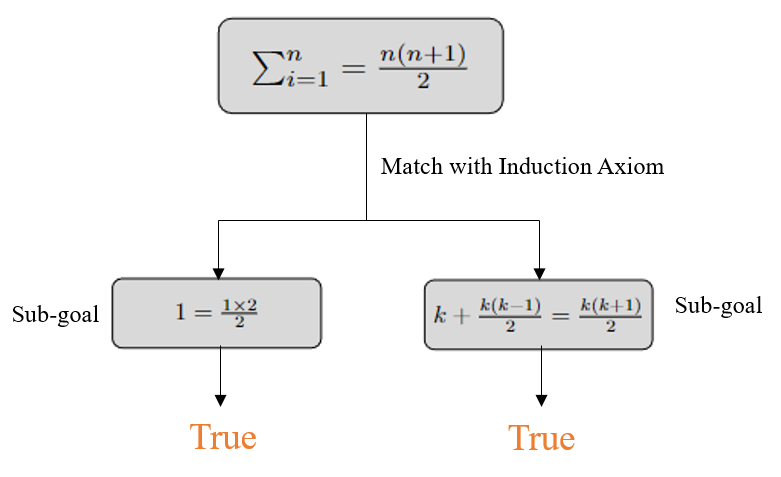}
    \end{minipage}
    \caption{Two examples of forward reasoning on the left and backward reasoning on the right.}

    \label{fig:fwd_bwd_reasoning}
\end{figure}

Theorem proving encompasses two types of reasoning: forward reasoning and backward reasoning. Forward reasoning involves identifying a pattern match between a particular theorem and the given conditions along with the universal variables, then deducing the conclusion. In contrast, backward reasoning works in the opposite direction, where the conclusion and variables are matched with a specific theorem, breaking down the main goal into smaller, more manageable subgoals. Both methods are essential in constructing and navigating the logical steps to establish the validity of complex mathematical theorems, as shown in Fig. \ref{fig:fwd_bwd_reasoning}.

\subsubsection{Challenges in Algebraic Reasoning}

There are two main challenges in reasoning within algebraic systems. The first is the infinite reasoning space within finite conditions, caused by the numerous possible expression trees and the vast search space for premises. This contrasts with solving Euclidean geometry problems, where a deduction fixed point exists with respect to a set of geometric rules or axioms. To address this issue, we consider only the current expression tree at each step of reasoning. The second challenge lies in pattern matching, which requires accurately identifying and applying relevant theorems to given sub-structures. For theorems with function-type variables, like Jensen's Inequality, pattern matching is more challenging and time-consuming. We provide heuristic functions to identify possible structures where Jensen's Inequality can be applied.





\newpage
\subsection{Theorems, Rules and Pattern Matching}
\subsubsection{Theorems, Methods and Transformation Rules}
Our deductive engine incorporates six well-known inequality theorems frequently used in mathematical Olympiads, several one-variable inequality scaling and solving methods, and dozens of algebraic transformation rules. The inequality theorems include the \textbf{Arithmetic Mean-Geometric Mean (AM-GM) inequality}, the \textbf{weighted AM-GM inequality}, \textbf{H\"older's inequality}, \textbf{Jensen's inequality}, \textbf{Schur's inequality}, and \textbf{M\"uirhead's theorem}. For simplicity, we have excluded some theorems that can be directly proved using these inequalities, such as the Geometric Mean-Harmonic Mean (GM-HM) inequality and the Cauchy-Schwarz inequality.

Here we list some frequently used methods and transformation rules:
\begin{itemize}
    \item \texttt{nodiv\_expr}: Multiply both sides to eliminate denominators
    \item \texttt{nomul\_expr}: Divide both sides by all factors
    \item \texttt{no\_sep\_denom}: Combine fractions on both sides
    \item \texttt{sep\_neg}: Move terms with negative coefficients to the other side of the inequality
    \item \texttt{zero\_side}: Subtract one side from the other to make one side equal to zero
    \item \texttt{no\_pow}: Remove roots at the second level from the top of the expression tree on both sides
    \item \texttt{try\_together\_l, try\_together\_r}: Combine fractions on the left or right side
    \item \texttt{try\_expand\_l, try\_expand\_r}: Expand expressions on the left or right side
    \item \texttt{all\_cyc\_mul\_expr}: Multiply both sides by a cyclically symmetric polynomial, with one of its generators on either the left or right side of the inequality (a generator is a term that, when cyclically permuted, generates the expression)
    \item \texttt{try\_factor\_both}: Factorize both sides
    \item \texttt{check\_one\_var}: Check if the solution of a one-variable inequality is contained in a given interval
    \item \texttt{check\_linear\_ctr}: Check if a one-variable expression can be applied with tangent line trick
    \item \texttt{find\_main\_fun}: For a cyclically symmetric expression, try to find a function that can match with Jensen's inequality as well as generate this expression
\end{itemize}

\subsubsection{Pattern Matching}
An important step in generating synthetic theorems is matching algebraic expressions with these theorems. We use the AM-GM inequality as an example to illustrate  pattern matching method as follows.

\begin{theorem}
    ({\bf AM-GM})
    For non-negative real numbers \(a_1, a_2, \ldots, a_n\),
    \[
    a_1 + a_2 + \cdots + a_n \geq n\sqrt[n]{a_1 a_2 \cdots a_n}
    \]
    with equality if and only if \(a_1 = a_2 = \cdots = a_n\).
\end{theorem}

Assuming all variables are non-negative, pattern matching for an algebraic expression with the AM-GM inequality (on the Left-Hand-Side) is explained in three steps:
\begin{enumerate}
    \item Traverse through the expression tree, and label a node with $1$ if the whole expression value increases as the value of the node increases, with $-1$ if the expression value decreases as the value of the node increases, and with \textit{None} if this cannot be determined.
    \item At each node labeled $1$ or $-1$ and calculated with an {\it Add} operation, find all non-negative sub-arguments of the node's expression and place them in nonneg\_set. Similarly, find all non-positive sub-arguments and place them in nonpos\_set.
    \item For the obtained sets nonneg\_set and nonpos\_set, we use the following method to match the mean inequalities:

    \begin{itemize}
        \item Arbitrarily partition each set into multiple subsets.
        \item the sum of the elements in each subset can be used as a variable to match the left side of the mean inequality.
        \item If a subset does not contribute to the inequality, it is excluded from the partition.
    \end{itemize}
    
    This process allows us to identify all possible mean inequalities that can be matched. We then replace the original sub-expressions in the expression tree with the transformed ones based on the matched inequalities. By doing so, a new inequality is derived according to the labels.

\end{enumerate}

\subsection{Details of Synthetic Data Generation}
Olympiad inequalities aim for not only difficulty but also conciseness and elegance, a principle also valued in modern mathematics. Although our deductive engine can generate various types of inequalities, we focus on cyclically symmetric inequalities in semi-definite systems that can be generated with a limited number of steps to avoid lengthy and chaotic expressions.

Initially, we generate thousands of premises as the starting points for data generation using Algorithm \ref{alg:random_generation}. For each generated premise, we run the data-generation algorithm described in the main paper. During this process, we discard inequalities for which equality does not hold or which do not have the desired form, and halt the generation after a maximum of 25 iterations of search. Utilizing 32 CPUs over an 8-hour period, the deductive engine produces 191,643 theorems. This demonstrates the engine's ability to efficiently generate a large number of high-quality inequality theorems, thereby addressing the bottleneck of lacking a high-quality dataset for learning-based provers.

\begin{algorithm}
\caption{Generating Initial Premises}
\label{alg:random_generation}
\begin{algorithmic}[Gen_exprs]
    \STATE \textbf{function} \texttt{GENERATE\_EXPRESSIONS}(\textit{variable\_list} $I$, \textit{loop\_limit} $N$)
    \begin{ALC@g}
        \STATE Initialize \textit{Results} and \textit{Basic\_Operations}
        \FOR{$i \leftarrow 1$ to $N$}
            \STATE Initialize \textit{New\_Expressions}
            \FOR{each pair $(a, b)$ in $I$ and each operation $f$ in \textit{Basic\_Operations}}
                \STATE Add $f(a, b)$ to \textit{New\_Expressions}
            \ENDFOR
            \STATE Add \textit{New\_Expressions} to $I$
        \ENDFOR
        \FOR{each expression \textit{expr} in $I$}
            \STATE Add cyclic summation of \textit{expr} to \textit{Results}
        \ENDFOR
        \STATE \textbf{return} \textit{Results}
\end{ALC@g}
\STATE \textbf{end function}
\end{algorithmic}
\end{algorithm}

\newpage
\section{Experiments and Analysis}
In this section, we provide details of our experiments and present the test results. We also include technical analysis of these results.
\subsection{Synthetic Dataset Statistics}
\begin{figure}[h]
    
    \centering
    \begin{minipage}{0.45\textwidth}
        \centering
        \includegraphics[width=\textwidth]{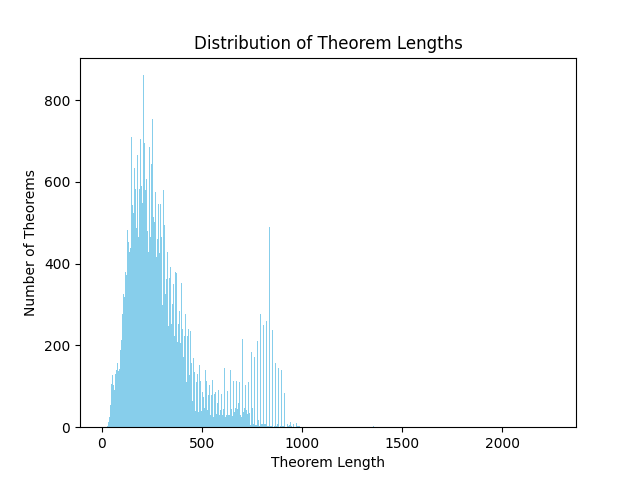}
    \end{minipage}
    \hfill
    \begin{minipage}{0.45\textwidth}
        \centering
        \includegraphics[width=\textwidth]{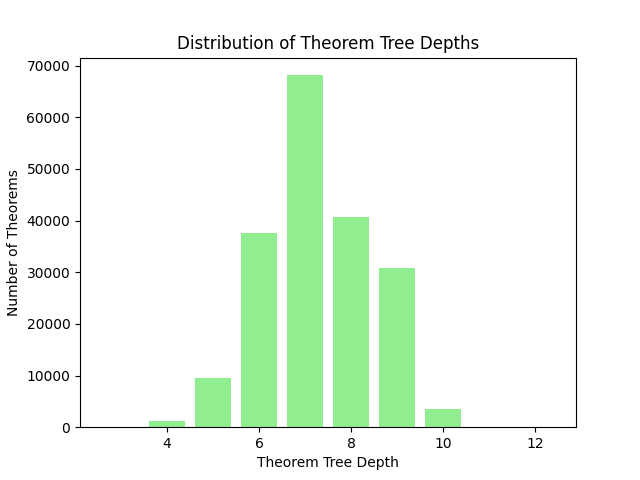}
    \end{minipage}
    \caption{Ditribution of lengths and tree-depths of synthetic theorems.}

    \label{fig:dis-len-dep}
\end{figure}

We conduct a statistical analysis on the synthetic dataset, focusing on inequality lengths (in string representation) and \textit{tree-depth} (the maximum expression tree height on both sides of an inequality), as depicted in Figure \ref{fig:dis-len-dep}. The distributions of lengths and tree-depth are related to the difficulty and search complexity. These distributions illustrate that our theorems range from simple to complex, reflecting a spectrum of difficulty levels in our dataset.

\subsection{Details of Value Curriculum Learning}
\label{v-learning}
The value network \( V_{\theta} \) comprises two main components: the pre-trained transformer encoder, Llemma-7b \citep{llemma}, followed by a \( 4096 \times 256 \times 1 \) multilayer perceptron that outputs a value in the interval \( (0, 1) \). Initially, AIPS successfully resolves 7 out of 20 problems from the test set using the pre-trained value network. 

The value network \( V_{\theta} \) functions as the heuristic in the best-first-search algorithm. It comprises two main components: the pre-trained transformer encoder, Llemma-7b, followed by a \( 4096 \times 256 \times 1 \) feedforward neural network that outputs a value in the interval \( (0, 1) \). Initially, AIPS successfully resolves 7 out of 20 problems from the test set using the pre-trained value network. 

The procedure of value curriculum learning is as follows. After successfully proving a theorem, each node along the proof path is relabeled with a value that is $\epsilon$ times its original value. For node that has been searched but is not part of the proof path, if its original label is $v$, the label of this node is updated at the end of this curriculum learning round according to the formula: $\max(m, v) \times \eta + 1-\eta$. Here $m$ represents the maximum value after modification among the proof path nodes. 
Subsequently, the relabeled nodes undergo 10 loops of fine-tuning training. We choose $\epsilon = 0.3 $ and $\eta = 0.7$.

Before the value curriculum learning process, we randomly select 1,200 theorems from the synthetic dataset, excluding theorems with an inference depth of less than 4. These theorems undergo a curriculum learning strategy tailored for the pre-trained model. We limit the time for solving each problem to 40 minites. 
During curriculum learning, the theorems are solved and trained in an ascending order, sorted first by tree-depth, then by theorem length. The first 150 problems are solved within a mere two hours. After four days of training, AIPS solves 892 out of the first 1,000 problems, with 887 successes in the first 950 theorems. Since it struggles to solve problems after the 950th theorem, we decide to halt the training process at the 1,000th problem.

\subsection{Performance Analysis During Curriculum Learning}
\label{perform-curr}
\begin{figure}[h]
        \centering
        \includegraphics[width=\textwidth]{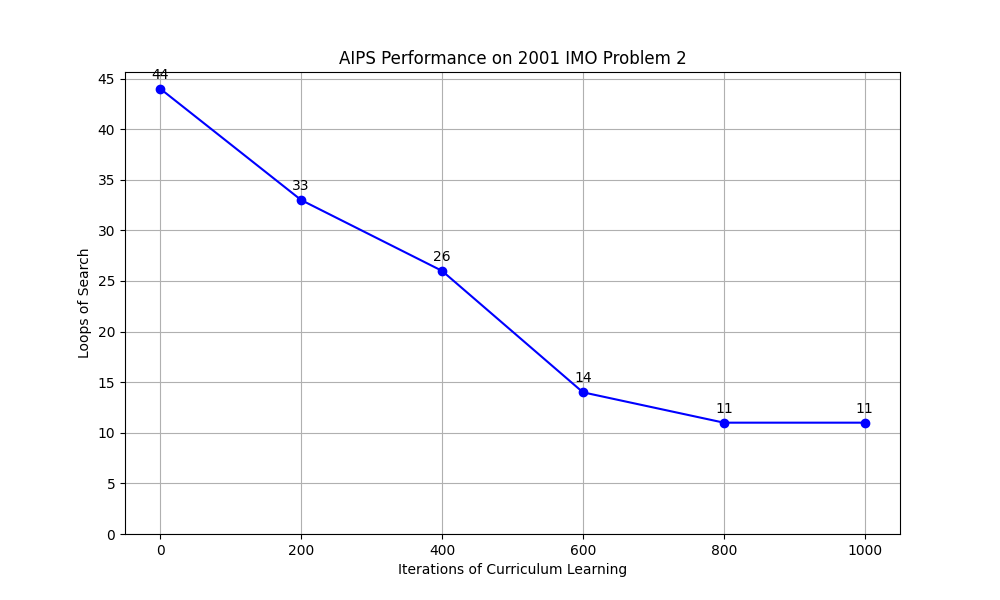}
        \caption{AIPS progressively finds the proof path more efficiently throughout the training process.}
        \label{fig:cl_search_loop}
\end{figure}

\begin{figure}[h]

        \centering
        \includegraphics[width=\textwidth]{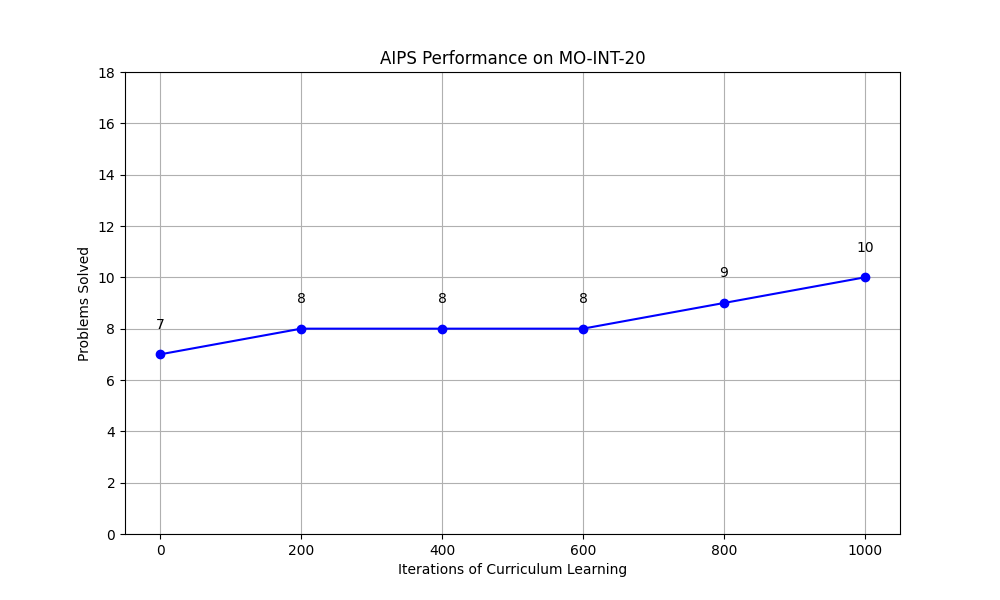}
        \caption{AIPS solves more problems with the increasing iterations of value curriculum learning.}

    \label{fig:cl_solved_problem}
\end{figure}

The extensive experiments verify that the value curriculum learning strategy is very effective. The number of search loops required to solve testing theorems decreases noticeably throughout the training process, enabling AIPS to successfully solve 10 out of 20 IMO-level inequality problems using an RTX-4090 GPU and a single CPU. Fig. \ref{fig:cl_search_loop} shows the decreasing number of search loops during curriculum learning on the 2001 IMO Problem 2, and Fig. \ref{fig:cl_solved_problem} shows the increasing number of solved problems during curriculum learning.

\subsection{Our Benchmark: Mathematical-Olympiad-INequality-Test-20}

We collect all ternary and quaternary algebraic inequality problems from IMO since 1990, some challenging problems from IMO shortlists and several national mathematical Olympiads, such as the USAMO, the USA National Team Selection Tests, the Polish/Korean/Japanese Mathematical Olympiad, all of which are of comparable difficulty to the IMO. The collected 20 problems provide a new challenging benchmark for the realm of automatic theorem proving, dubbed as MO-INT-20 (Math-Olympiad-INequality-Test-20). 
The details of these 20 problems are as follows.

\begin{itemize}
    \item \textbf{Problem 1 (IMO 1990 Shortlist):} \\
    For \(a > 0, b > 0, c > 0, d > 0\) such that \(a \cdot b + b \cdot c + c \cdot d + d \cdot a = 1\), show that:
    \[
    \frac{a^3}{b+c+d} + \frac{b^3}{c+d+a} + \frac{c^3}{d+a+b} + \frac{d^3}{a+b+c} \geq \frac{1}{3}
    \]

    \item \textbf{Problem 2 (IMO 1993 Shortlist):} \\
    For \(a > 0, b > 0, c > 0, d > 0\), show that:
    \[
    \frac{a}{b + 2c + 3d} + \frac{b}{3a + c + 2d} + \frac{c}{2a + 3b + d} + \frac{d}{a + 2b + 3c} \geq \frac{2}{3}
    \]

    \item \textbf{Problem 3 (IMO 1995 P2):} \\
    For \(a > 0, b > 0, c > 0\) such that \(a \cdot b \cdot c = 1\), show that:
    \[
    \frac{1}{c^3(a + b)} + \frac{1}{b^3(a + c)} + \frac{1}{a^3(b + c)} \geq \frac{3}{2}
    \]

    \item \textbf{Problem 4 (IMO 1996 Shortlist):} \\
    For \(a > 0, b > 0, c > 0\) such that \(a \cdot b \cdot c = 1\), show that:
    \[
    \frac{a \cdot b}{a^5 + a \cdot b + b^5} + \frac{a \cdot c}{a^5 + a \cdot c + c^5} + \frac{b \cdot c}{b^5 + b \cdot c + c^5} \leq 1
    \]

    \item \textbf{Problem 5 (USAMO 1997 P5):} \\
    For \(a > 0, b > 0, c > 0\), show that:
    \[
    \frac{1}{a^3 + b^3 + a \cdot b \cdot c} + \frac{1}{b^3 + c^3 + a \cdot b \cdot c} + \frac{1}{c^3 + a^3 + a \cdot b \cdot c} \leq \frac{1}{a \cdot b \cdot c}
    \]

    \item \textbf{Problem 6 (IMO 1998 Shortlist A3):} \\
    For \(a > 0, b > 0, c > 0\) such that \(a \cdot b \cdot c = 1\), show that:
    \[
    \frac{a^3}{(1 + b)(1 + c)} + \frac{b^3}{(1 + c)(1 + a)} + \frac{c^3}{(1 + a)(1 + b)} \geq \frac{3}{4}
    \]

    \item \textbf{Problem 7 (IMO 2000 P2):} \\
    For \(a > 0, b > 0, c > 0\) such that \(a \cdot b \cdot c = 1\), show that:
    \[
    (a - 1 + \frac{1}{b})(b - 1 + \frac{1}{c})(c - 1 + \frac{1}{a}) \leq 1
    \]

    \item \textbf{Problem 8 (IMO 2001 P2):} \\
    For \(a > 0, b > 0, c > 0\), show that:
    \[
    \frac{a}{\sqrt{a^2 + 8bc}} + \frac{b}{\sqrt{8ac + b^2}} + \frac{c}{\sqrt{8ab + c^2}} \geq 1
    \]

    \item \textbf{Problem 9 (USAMO 2003 P5):} \\
    For \(a > 0, b > 0, c > 0\), show that:
    \[
    \frac{(a + b + 2c)^2}{2c^2 + (a + b)^2} + \frac{(a + 2b + c)^2}{2b^2 + (a + c)^2} + \frac{(2a + b + c)^2}{2a^2 + (b + c)^2} \leq 8
    \]

    \item \textbf{Problem 10 (Poland 2004):} \\
    For \(a > 0, b > 0, c > 0, d > 0\), show that:
    \[
    \frac{a}{(a^3 + 63bcd)^{\frac{1}{3}}} + \frac{b}{(63acd + b^3)^{\frac{1}{3}}} + \frac{c}{(63abd + c^3)^{\frac{1}{3}}} + \frac{d}{(63abc + d^3)^{\frac{1}{3}}} \geq 1
    \]

    \item \textbf{Problem 11 (IMO 2004 Shortlist A5):} \\
    For \(a > 0, b > 0, c > 0\) such that \(a \cdot b + b \cdot c + c \cdot a = 1\), show that:
    \[
    \left( \frac{1}{a} + 6b \right)^{\frac{1}{3}} + \left( \frac{1}{b} + 6c \right)^{\frac{1}{3}} + \left( \frac{1}{c} + 6a \right)^{\frac{1}{3}} \leq \frac{1}{a \cdot b \cdot c}
    \]

    \item \textbf{Problem 12 (IMO 2006 P3):} \\
    Given real numbers \(a,b,c\), show that:
    \[
    |ab(a^2-b^2)+bc(b^2-c^2)+ca(c^2-a^2))| \leq \frac{9}{16\sqrt{2}}(a^2+b^2+c^2)^2
    \]

    \item \textbf{Problem 13 (IMO 2009 Shortlist):} \\
    For \(a > 0, b > 0, c > 0\) such that \(\frac{1}{a} + \frac{1}{b} + \frac{1}{c} = a + b + c\), show that:
    \[
    (2a + b + c)^{-2} + (a + 2b + c)^{-2} + (a + b + 2c)^{-2} \leq \frac{3}{16}
    \]

    \item \textbf{Problem 14 (USA IMO Team Selection 2010 P2):} \\
    For \(a > 0, b > 0, c > 0\) such that \(a \cdot b \cdot c = 1\), show that:
    \[
    \frac{1}{c^5(a + 2b)^2} + \frac{1}{b^5(2a + c)^2} + \frac{1}{a^5(b + 2c)^2} \geq \frac{1}{3}
    \]

    \item \textbf{Problem 15 (USAMO 2011 P1):} \\
    For \(a > 0, b > 0, c > 0\) such that \(a^2 + b^2 + c^2 + (a + b + c)^2 \leq 4\), show that:
    \[
    \frac{a \cdot b + 1}{(a + b)^2} + \frac{b \cdot c + 1}{(b + c)^2} + \frac{c \cdot a + 1}{(c + a)^2} \geq 3
    \]

    \item \textbf{Problem 16 (Korea 2011 P4):} \\
    For \(a \geq 0, b \geq 0, c \geq 0\) such that \(a + b + c = 1\), show that:
    \[
    \frac{1}{a^2 - 4a + 9} + \frac{1}{b^2 - 4b + 9} + \frac{1}{c^2 - 4c + 9} \leq \frac{7}{18}
    \]

    \item \textbf{Problem 17 (USAMO 2012):} \\
    For \(a > 0, b > 0, c > 0\), show that:
    \[
    \frac{b^3 + 3c^3}{5b + c} + \frac{a^3 + 3b^3}{5a + b} + \frac{3a^3 + c^3}{a + 5c} \geq \frac{2}{3}(a^2 + b^2 + c^2)
    \]

    \item \textbf{Problem 18 (Japan 2014 P5):} \\
    For \(a \geq 0, b \geq 0, c \geq 0\) such that \(a + b + c = 1\), show that:
    \[
    \frac{a}{9bc + 4(b - c)^2 + 1} + \frac{b}{9ac + 4(-a + c)^2 + 1} + \frac{c}{9ab + 4(a - b)^2 + 1} \geq \frac{1}{2}
    \]

    \item \textbf{Problem 19 (USAMO 2017 P6):} \\
    For \(a \geq 0, b \geq 0, c \geq 0, d \geq 0\) such that \(a + b + c + d = 4\), show that:
    \[
    \frac{a}{b^3 + 4} + \frac{b}{c^3 + 4} + \frac{c}{d^3 + 4} + \frac{d}{a^3 + 4} \geq \frac{2}{3}
    \]


    \item \textbf{Problem 20 (IMO 2020 P2):} \\
    For \(a \geq b, b \geq c, c \geq d, d > 0\) such that \(a + b + c + d = 1\), show that:
    \[
    (a + 2b + 3c + 4d)a^a b^b c^c d^d < 1
    \]
\end{itemize}

\subsection{Details of Comparison Methods and Testing Results}
\subsubsection{Monte-Carlo Tree Search}
We evaluate the performance of Monte-Carlo Tree Search (MCTS). Compared to games like Go or chess, theorem proving can have an extremely large or even infinite action space, since applying each theorem or axiom usually comes with a set of parameters. Therefore, a direct application of MCTS to our problems is infeasible. To address this, we need to modify the MCTS algorithm.

First, we place a restriction on our action space: at each state, we sample all possible actions generated from the current proof state, then sort them according to a tree-depth heuristic function, which evaluates the difficulty of the proof state after applying them, and pick the first $k$ proof states (we choose $k = 5$). During the selection step in MCTS, we apply the Upper Confidence Bounds algorithm,
\[
\text{SelectedAction} = \text{Argmax}_i\left(v_i + C \cdot \sqrt{\frac{\ln(N)}{n_i}}\right)
\]
Here, $n_i$ is the number of visits of the $i$-th action (proof state), $N$ is the number of visits of the current state, $C$ is an adjustable hyperparameter (we choose $C=0.3\sqrt{2}$), and $v_i$ is the value of the state after applying the $i$-th action.

We use two methods to compute $v_i$: either it is directly given by the heuristic evaluation function, or we execute a best-first search to a constant depth $d$ ($d = 2$), and take the value of the current best state as $v_i$.

\subsubsection{Testing Examples from Large Language Models}
\label{llm-eg}
Large Language Models (LLMs) currently struggle to solve IMO-level inequalities through the generation of natural language proofs. In our evaluation of 4 LLMs—GPT-4, GPT-4 Turbo, Gemini 1.5 Pro, and Llemma-7b—only Gemini 1.5 Pro successfully generates a correct solution to USAMO-1997-P5. Proofs generated by GPT-4 and GPT-4 Turbo contain some logical errors, while Llemma-7b rarely produces meaningful reasoning sentences.
We present the results of first three LLMs below, including the only correct proof by Gemini 1.5 Pro to USAMO-1997-P5,  and the incorrect proofs by using these three LLMs to IMO-1995-P2. 

\begin{figure}[!h]
\vspace{-0.3in}
\begin{tcolorbox}[colback=yellow!8!]
\begin{small}
\textbf{Statement}: \,(USAMO 1997 P5)\, Given three positive real numbers $a, b, c$, show that 
\[\frac{1}{a^3+b^3+abc}+\frac{1}{b^3+c^3+abc}+\frac{1}{c^3+a^3+abc}\leq \frac{1}{abc}\]

\noindent\makebox[\linewidth]{\rule{\textwidth}{1pt}} 
\textbf{Correct proof}: \par
\vspace{5pt}
\setlength\parindent{8pt}
1. Factorization: 
Recall the factorization of the sum of cubes:
\[x^3+y^3=(x+y)(x^2-xy+y^2)\]
\hspace{\parindent}
2. Apply AM-HM:
For non-negative numbers $x$ and $y$, the AM-HM inequality states:
\[\frac{x+y}{2}\geq \frac{2}{1/x+1/y}\]
\hspace{\parindent}

Rearranging the inequality:
\[x^2+y^2-xy\geq xy\]
\hspace{\parindent}
3. Manipulate the given inequality:
Lets focus on the first term and use the factorization:
\[\frac{1}{a^3+b^3+abc}=\frac{1}{(a+b)(a^2-ab+b^2)+abc}\]
\hspace{\parindent}

Since $a^2-ab+b^2\geq ab$ (from step 2), we have:
\[\frac{1}{(a+b)(a^2-ab+b^2)+abc}\leq \frac{1}{(a+b)ab+abc}=\frac{1}{(a+b+c)ab}\]
\hspace{\parindent}
4. Apply similar reasoning to other terms:
Similarly, for the second and third terms:
\[\frac{1}{b^3+c^3+abc}\leq \frac{1}{(a+b+c)bc}\]
\[\frac{1}{c^3+a^3+abc}\leq \frac{1}{(a+b+c)ca}\]
\hspace{\parindent}
5. Combine the inequalities:
Adding the three inequalities we just derived and simplify the right hand side:
\[\frac{1}{a^3+b^3+abc}+\frac{1}{b^3+c^3+abc}+\frac{1}{c^3+a^3+abc}\leq \frac{1}{abc}\]
\hspace{\parindent}

This completes the proof.\par
\end{small}
\end{tcolorbox}
\caption{
   Correct proof generated by Gemini 1.5 Pro.}
\label{correct}
\end{figure}

\begin{figure}[t]
\vspace{-0.3in}
\begin{tcolorbox}[colback=yellow!8!]
\begin{small}
\label{gemini}
\textbf{Statement}: \,(IMO 1995 P2)\, Given positive real numbers $a,b,c$ such that $abc=1$, show that 
\[\frac{1}{c^3(a+b)}+\frac{1}{b^3(a+c)}+\frac{1}{a^3(b+c)}\geq 3/2\]
\noindent\makebox[\linewidth]{\rule{\textwidth}{1pt}}

\textbf{Incorrect proof:} (Generated by Gemini 1.5 Pro)
\setlength\parindent{8pt}\par
Transformation: First, let's simplify the given expression by multiplying the numerator and denominator of each term by abc (which is equal to 1 due to the condition abc = 1):
\[\frac{1}{c^3(a + b)} + \frac{1}{b^3(a + c)} + \frac{1}{a^3(b + c)} 
= \frac{ab}{c^2(a + b)} + \frac{ac}{b^2(a + c)} + \frac{bc}{a^2(b + c)}\]\par
Applying AM-HM: Now, consider the following positive numbers:
\[x_1 = c^2(a + b),
x_2 = b^2 (a + c),
x_3 = a^2 (b + c)\]\par
Applying AM-HM inequality to these:
\[\frac{c^2(a+b) + b^2(a+c) + a^2(b+c)}{3}  \geq  \frac{3}{\frac{1}{c^3(a + b)} + \frac{1}{b^3(a + c)} + \frac{1}{a^3(b + c)} }\]\par
Simplifying: Expanding and simplifying the left side:
\[\frac{a^2b + ab^2 + ac^2 + a^2c + b^2c + bc^2}{3} \geq \frac{3}{\frac{ab}{c^2(a + b)} + \frac{ac}{b^2(a + c)} + \frac{bc}{a^2(b + c)}}\]

\textcolor{blue}{Since abc = 1, we can substitute and further simplify:
\[\frac{ab/c + ac/b + bc/a}{3} \geq \frac{3}{\frac{1}{c^3(a + b)} + \frac{1}{b^3(a + c)} + \frac{1}{a^3(b + c)}}\]} 

\textcolor{blue}{Final Step: Invert both sides of the inequality (remembering to flip the inequality sign):
\[\frac{1}{c^3(a + b)} + \frac{1}{b^3(a + c)} + \frac{1}{a^3(b + c)} \geq 3/2\]}

This proves the required inequality.
\end{small}
\end{tcolorbox}
\caption{
    Incorrect proof generated by Gemini 1.5 Pro. The errors are labeled in blue.}

\label{correct}
\end{figure}

\begin{figure}[t]\label{GPT4-turbo}
\vspace{-0.3in}
\begin{tcolorbox}[colback=yellow!8!]
\begin{small}
\textbf{Statement}: \,(IMO 1995 P2)\, Given positive real numbers $a,b,c$ such that $abc=1$, show that 
\[\frac{1}{c^3(a+b)}+\frac{1}{b^3(a+c)}+\frac{1}{a^3(b+c)}\geq 3/2\]
\noindent\makebox[\linewidth]{\rule{\textwidth}{1pt}}
\textbf{Incorrect proof:} (Generated by ChatGPT4 Turbo)
\setlength\parindent{8pt}\par
To prove this inequality, we can apply the AM-GM (Arithmetic Mean - Geometric Mean) inequality as follows:\par
Given $abc = 1$:\par
By applying the AM-GM inequality:
\[a/b + b/a \geq 2,
b/c + c/b \geq 2,
c/a + a/c \geq 2\]
\par
Summing these inequalities:
\[(a/b + b/a) + (b/c + c/b) + (c/a + a/c) \geq 6\]
\par
Rearrange and factor:
\[a^2/b^2 + b^2/c^2 + c^2/a^2 + b^2/a^2 + c^2/b^2 + a^2/c^2 \geq 6\]
\par
Now consider the expressions under the original inequality:
\[\frac{1}{a^3(b+c)} = \frac{1}{a^3(b+c)}\]
\par
Since abc = 1, we can write this as:
\[= \frac{b^3c^3}{a^3(b+c)}
= \frac{(bc)^3}{a^3(b+c)}\]
\par
\textcolor{blue}{By AM-GM, $b+c \geq 2\sqrt{bc}$. Therefore:
\[\frac{1}{a^3(b+c)} \geq \frac{(bc)^3}{2a^3\sqrt{bc}}
= \frac{b^2c^2}{2a^3\sqrt{bc}} \cdot \sqrt{bc}
= \frac{b^{5/2}c^{5/2}}{2a^3}\]}
\par
We can apply similar transformations to the other fractions:
\[\frac{1}{b^3(c+a)} \geq \frac{a^{5/2}c^{5/2}}{2b^3},
\frac{1}{c^3(a+b)} \geq \frac{a^{5/2}b^{5/2}}{2c^3}\]
\par
Adding these up:
\[\frac{1}{a^3(b+c)} + \frac{1}{b^3(c+a)} + \frac{1}{c^3(a+b)} \geq \frac{b^{5/2}c^{5/2}}{2a^3} + \frac{a^{5/2}c^{5/2}}{2b^3} + \frac{a^{5/2}b^{5/2}}{2c^3}\]
\par
Using our earlier inequality
\[(a^2/b^2 + b^2/a^2 + b^2/c^2 + c^2/b^2 + c^2/a^2 + a^2/c^2 \geq 6)\]
we can apply the 
generalized AM-GM inequality:
\[\frac{1}{a^3(b+c)} + \frac{1}{b^3(c+a)} + \frac{1}{c^3(a+b)} \geq \frac{1}{2}\cdot (\frac{b^{5/2}c^{5/2}}{a^3} + \frac{a^{5/2}c^{5/2}}{b^3} + \frac{a^{5/2}b^{5/2}}{c^3})\]
\par
Each of these terms is equal to 1 by our substitution using abc = 1, simplifying this expression:
\[\geq 1/2 + 1/2 + 1/2 = 3/2\]
\par
Therefore, we have proved:
\[\frac{1}{a^3(b+c)} + \frac{1}{b^3(c+a)}+\frac{1}{c^3(a+b)} \geq 3/2\]
\end{small}
\end{tcolorbox}
\caption{
 Incorrect proof generated by ChatGPT4 Turbo. The errors are labeled in blue.
}
\end{figure}

\begin{figure}[t]\label{GPT4}
\vspace{-0.3in}
\begin{tcolorbox}[colback=yellow!8!]
\begin{small}
\textbf{Statement}: \,(IMO 1995 P2)\, Given positive real numbers $a,b,c$ such that $abc=1$, show that 
\[\frac{1}{c^3(a+b)}+\frac{1}{b^3(a+c)}+\frac{1}{a^3(b+c)}\geq 3/2\]
\noindent\makebox[\linewidth]{\rule{\textwidth}{1pt}}
\textbf{Incorrect proof:} (Generated by ChatGPT 4)\par
\vspace{5pt}
\setlength\parindent{8pt}
The Arithmetic Mean - Geometric Mean (AM-GM) Inequality states:
\[\frac{x_1+x_2+\dots +x_n}{n}\geq \sqrt[n]{x_1\cdot x_2\dots x_n}\]
\hspace{\parindent}
with equality if and only if $x_1 = x_2\dots = x_n$. Let's consider using AM-GM on the denominators.\par
Analyzing Each Fraction\par
For $\frac{1}{c^3(a+b)}$: Since $abc=1$, we have $c^3 = \frac{1}{a^3b^3}$. Hence, 
\[\frac{1}{c^3(a+b)} = \frac{a^3b^3}{a+b}\]
\hspace{\parindent}
Similiarly:
\[\frac{1}{b^3(a+c)} = \frac{a^3c^3}{a+c},\,\frac{1}{a^3(b+c)} = \frac{b^3c^3}{b+c}\]
\hspace{\parindent}
Here each fraction can be analyzed by using AM-GM on the denominators.
\[a+b\geq 2\sqrt{ab},b+c\geq 2\sqrt{bc},c+a\geq 2\sqrt{ca}\]
\hspace{\parindent}
\textcolor{blue}{Given $abc=1$, we also have $\sqrt{ab}=\frac{c}{\sqrt{abc}}=c$ and similarly for others.} Thus, 
\[a+b\geq 2c, b+c\geq 2a, c+a\geq 2b\]
\hspace{\parindent}
Substituting back:
\[\frac{a^3b^3}{a+b}\geq\frac{a^3b^3}{2c},\frac{b^3c^3}{b+c}\geq\frac{b^3c^3}{2a},\frac{c^3a^3}{c+a}\geq\frac{c^3a^3}{2b}\]
\hspace{\parindent}
\textcolor{blue}{Since $a^3b^3=c^3,b^3c^3=a^3,c^3a^3=b^3$ from $abc=1$,}
\[\frac{a^3b^3}{2c}=\frac{c^2}{2},\frac{b^3c^3}{2a}=\frac{a^2}{2},\frac{c^3a^3}{2b}=\frac{b^2}{2}\]
\hspace{\parindent}
Adding these,
\[\frac{a^2}{2}+\frac{b^2}{2}+\frac{c^2}{2}=\frac{a^2+b^2+c^2}{2}\]
\hspace{\parindent}
To conclude, we need the final step:
\[a^2+b^2+c^2\geq 3\]
\hspace{\parindent}
which holds by AM-GM as $a^2+b^2+c^2\geq 3\sqrt[3]{a^2b^2c^2}=3$. This completes the proof.
\end{small}
\end{tcolorbox}
\caption{
Incorrect proof generated by ChatGPT 4. The errors are labeled in blue.
}
\end{figure}
\clearpage
\subsubsection{Testing Examples from LeanCopilot}
\label{test-res-lean}
We also test LeanCopilot (developed from LeanDojo \citet{leandojo}), the current open-source state-of-the-art theorem prover based on Lean. It solves none of the 20 problems, all resulting in time-limited feedback. We present some results of the tactic search provided by LeanCopilot in Figure \ref{lean-copilot}. It can be seen that the prover tends to rely heavily on built-in tactics and struggles to predict complex premises.

\vspace{24pt}
\begin{figure}[!h]
\vspace{-0.1in}
\begin{tcolorbox}[colback=yellow!8!]
\begin{footnotesize}
\begin{lstlisting}
USAMO-1997-P5:

theorem (a b c:R)(h0:a>0)(h1:b>0)(h2:c>0):
1/(a^3+b^3+a*b*c)+1/(b^3+c^3+a*b*c)+1/(c^3+a^3+a*b*c)<=1/(a*b*c)

Try these:
nth_rw 1 [\l mul_one (a*b*c)]  #Replace abc by 1*abc
ring_nf                        #Simplify by ring axiom
field_simp                     #Simplify by field axiom                      
refine' le_of_eq _             #Proving inequality by equality
rw [one_div]                   #Replace 1/x by x^(-1)
nth_rw 3 [\l mul_one(a*b*c)]   #Replace abc by 1*abc
rw [le_div_iff']               #Multiply abc on both sides
\end{lstlisting}
\noindent\makebox[\linewidth]{\rule{\textwidth}{1pt}} 

\begin{lstlisting}
IMO-1995-P2:

theorem (a b c:R)(h0:a>0)(h1:b>0)(h2:c>0)(h3: a*b*c=1):
1/(c^3*3*(a+b))+1/(b^3*3*(a+c))+1/(a^3*3*(b+c)) >= 3/2 

Try these:
refine' le_of_eq _            #Proving inequality by equality
norm_num                      #Normalize numerical expressions
rw [\l h3]                    #Replace 1 by abc
field_simp                    #Simplify by field axiom 
ring_nf                       #Simplify by ring axiom
field_simp [h1, h2]           #Simplify by field axiom + h1,h2                         
push_cast                     #Move certain coercions inward
\end{lstlisting}

\end{footnotesize}
\end{tcolorbox}
\caption{
\small
\textbf{Tactics suggested by LeanCopilot to two problems, namely USAMO-1997-P5 and IMO-1995-P2.} 
}
\label{lean-copilot}
\end{figure}

\subsubsection{10 Problems Solved by Our AIPS}
When proving an inequality, AIPS first homogenizes both sides using the given conditions if the inequality is not already homogenized, thereby obtaining a new inequality. It then performs mixed reasoning on the new inequality to complete the proof. We present the proofs for the 10 problems solved by our AIPS as follows.\par

\begin{figure}[t]
\vspace{-0.3in}
\begin{tcolorbox}[colback=orange!8!]
\begin{small}
\textbf{1. Solution to IMO-1990-Shortlist Problem}\\
By \texttt{<function try\_homo>}, It is equivalent to prove
$$\frac{a^{3}}{b + c + d} + \frac{b^{3}}{a + c + d} + \frac{c^{3}}{a + b + d} + \frac{d^{3}}{a + b + c} \geq \frac{a b}{3} + \frac{a d}{3} + \frac{b c}{3} + \frac{c d}{3}$$
by \texttt{<function check\_AM\_GM\_Mul2>}, it remains to prove
$$\frac{a^{2}}{3} + \frac{b^{2}}{3} + \frac{c^{2}}{3} + \frac{d^{2}}{3} \leq \frac{a^{3}}{b + c + d} + \frac{b^{3}}{a + c + d} + \frac{c^{3}}{a + b + d} + \frac{d^{3}}{a + b + c}$$
by \texttt{<function try\_together\_l>}, it remains to prove
$$\frac{a^{2} + b^{2} + c^{2} + d^{2}}{3} \leq \frac{a^{3}}{b + c + d} + \frac{b^{3}}{a + c + d} + \frac{c^{3}}{a + b + d} + \frac{d^{3}}{a + b + c}$$
we use H\"older's inequality: 
\[(a^2 + b^2 + c^2 + d^2)^2 \leq\]
\[(a(b + c + d) + b(a + c + d) + c(a + b + d) + d(a + b + c))\times\]
\[(a^3/(b + c + d) + b^3/(a + c + d) + c^3/(a + b + d) + d^3/(a + b + c)).\] 
It remains to prove
$$\frac{a^{2} + b^{2} + c^{2} + d^{2}}{3} \leq \frac{\left(a^{2} + b^{2} + c^{2} + d^{2}\right)^{2}}{a \left(b + c + d\right) + b \left(a + c + d\right) + c \left(a + b + d\right) + d \left(a + b + c\right)}$$
by \texttt{<function all\_cyc\_mul\_expr>}, it remains to prove
$$\frac{1}{3} \leq \frac{a^{2} + b^{2} + c^{2} + d^{2}}{a \left(b + c + d\right) + b \left(a + c + d\right) + c \left(a + b + d\right) + d \left(a + b + c\right)}$$
For $f(x) = x^2$, $f''(x)>0 \text{ for } 0<x$.
we use Jensen's inequality: $$4(a/4 + b/4 + c/4 + d/4)^2 \leq a^2 + b^2 + c^2 + d^2,$$ it remains to prove
$$\frac{1}{3} \leq \frac{4 \left(\frac{a}{4} + \frac{b}{4} + \frac{c}{4} + \frac{d}{4}\right)^{2}}{a \left(b + c + d\right) + b \left(a + c + d\right) + c \left(a + b + d\right) + d \left(a + b + c\right)}$$
For $f(x) = x(a+b+c+d-x)$, $f''(x)<0 \text{ for } 0<x<a+b+c+d$, we use Jensen's inequality:
\[a(b + c + d) + b(a + c + d) + c(a + b + d) + d(a + b + c) \leq\]
\[4(a/4 + b/4 + c/4 + d/4)(3a/4 + 3b/4 + 3c/4 + 3d/4)\]
it remains to prove
$$\frac{1}{3} \leq \frac{\frac{a}{4} + \frac{b}{4} + \frac{c}{4} + \frac{d}{4}}{\frac{3 a}{4} + \frac{3 b}{4} + \frac{3 c}{4} + \frac{3 d}{4}}$$
by \texttt{<function try\_simp\_r>}, this is true!
\end{small}
\end{tcolorbox}
%
\end{figure}
\begin{figure}[t]
\vspace{-0.3in}
\begin{tcolorbox}[colback=orange!8!]
\begin{small}
\textbf{2. Solution to IMO-1993-Shortlist problem.}\\
To prove
$$\frac{a}{b + 2 c + 3 d} + \frac{b}{3 a + c + 2 d} + \frac{c}{2 a + 3 b + d} + \frac{d}{a + 2 b + 3 c} \geq \frac{2}{3}$$
we use H\"older's inequality: 
\[(a + b + c + d)^2 \leq 
(\dfrac{a}{b + 2c + 3d}) + \dfrac{b}{3a+c+2d} + \dfrac{c}{2a+3b+d} + \dfrac{d}{a+2b+3c})\times\]
\[(a(b + 2c + 3d) + b(3a + c + 2d) + c(2a + 3b + d) + d(a + 2b + 3c)).\]
\\It remains to prove
$$\frac{2}{3} \leq \frac{\left(a + b + c + d\right)^{2}}{4 a b + 4 a c + 4 a d + 4 b c + 4 b d + 4 c d}$$
by \texttt{<function all\_cyc\_mul\_expr>}, it remains to prove
$$\frac{2}{3 \left(a + b + c + d\right)^{2}} \leq \frac{1}{4 a b + 4 a c + 4 a d + 4 b c + 4 b d + 4 c d}$$
by \texttt{<function try\_expand\_l>}, it remains to prove
\[\frac{2}{3 a^{2} + 6 a b + 6 a c + 6 a d + 3 b^{2} + 6 b c + 6 b d + 3 c^{2} + 6 c d + 3 d^{2}} \leq\] \[\frac{1}{4 a b + 4 a c + 4 a d + 4 b c + 4 b d + 4 c d}\]
by \texttt{<function nodiv\_expr>}, it remains to prove
$$8 a b + 8 a c + 8 a d + 8 b c + 8 b d + 8 c d \leq 3 a^{2} + 6 a b + 6 a c + 6 a d + 3 b^{2} + 6 b c + 6 b d + 3 c^{2} + 6 c d + 3 d^{2}$$
by \texttt{<function zero\_side>}, it remains to prove
$$0 \leq 3 a^{2} - 2 a b - 2 a c - 2 a d + 3 b^{2} - 2 b c - 2 b d + 3 c^{2} - 2 c d + 3 d^{2}$$
by \texttt{<function check\_AM\_GM\_Mul2>}, it remains to prove
$$0 \leq 2 a^{2} - 2 a b - 2 a d + 2 b^{2} - 2 b c + 2 c^{2} - 2 c d + 2 d^{2}$$
by \texttt{<function check\_AM\_GM\_Mul2>}, this is true!
\end{small}
\end{tcolorbox}
%
\end{figure}\par

\begin{figure}[h]
\vspace{-0.3in}
\begin{tcolorbox}[colback=orange!8!]
\begin{small}
\textbf{3. Solution to IMO-1995-P2}\\
By \texttt{<function try\_homo>}, it is equivalent to prove
$$\dfrac{a^2b^2}{c(a+b)}+\dfrac{b^2c^2}{a(b+c)}+\dfrac{a^2c^2}{b(a+c)}\geq \dfrac{3a^\frac{2}{3}b^\frac{2}{3}c^\frac{2}{3}}{2}$$
We use H\"older's inequality:
$$\dfrac{ab+bc+ca}{2}\leq (c(a+b)+a(b+c)+b(c+a))(\dfrac{a^2b^2}{c(a+b)}+\dfrac{b^2c^2}{a(b+c)}+\dfrac{a^2c^2}{b(a+c)}).$$
It remains to prove $$\dfrac{3a^\frac{2}{3}b^\frac{2}{3}c^\frac{2}{3}}{2}\leq \dfrac{ab+bc+ca}{2}$$
by \texttt{<function check\_AM\_GM>}, this is true!
\end{small}
\end{tcolorbox}
%

\end{figure}
\begin{figure}[t]
\vspace{-0.3in}
\begin{tcolorbox}[colback=orange!8!]
\begin{small}
\textbf{4. Solution to USAMO-1997-P5.}\\
To prove
$$\frac{1}{a b c + b^{3} + c^{3}} + \frac{1}{a^{3} + a b c + c^{3}} + \frac{1}{a^{3} + a b c + b^{3}} \leq \frac{1}{a b c}$$
by \texttt{<function check\_SimpMuirhead>}, it remains to prove
$$\frac{1}{a b c + b^{2} c + b c^{2}} + \frac{1}{a^{2} c + a b c + a c^{2}} + \frac{1}{a^{2} b + a b^{2} + a b c} \leq \frac{1}{a b c}$$
by \texttt{<function try\_together\_l>}, this is true!
\end{small}
\end{tcolorbox}
%
\end{figure}

\begin{figure}[h]
\vspace{-0.3in}
\begin{tcolorbox}[colback=orange!8!]
\begin{small}
\textbf{5. Solution to 2001-IMO-P2.}\\
To prove
$$
\frac{a}{\sqrt{a^{2} + 8bc}} + \frac{b}{\sqrt{8ac + b^{2}}} + \frac{c}{\sqrt{8ab + c^{2}}} \geq 1,
$$
we use H\"older's inequality:
\[(a + b + c)^3 \leq \]
\[\left( \frac{a}{\sqrt{a^2 + 8bc}} + \frac{b}{\sqrt{8ac + b^2}} + \frac{c}{\sqrt{8ab + c^2}} \right)^2 \left(a(a^2 + 8bc) + b(8ac + b^2) + c(8ab + c^2)\right).\]
It remains to prove
$$
1 \leq \frac{\left(a + b + c\right)^{\frac{3}{2}}}{\sqrt{a^{3} + 24abc + b^{3} + c^{3}}}.
$$
by \texttt{<function all\_cyc\_mul\_expr>}, it remains to prove
$$
\sqrt{a^{3} + 24abc + b^{3} + c^{3}} \leq \left(a + b + c\right)^{\frac{3}{2}}
$$
by \texttt{<function no\_pow>}, it remains to prove
$$
a^{3} + 24abc + b^{3} + c^{3} \leq \left(a + b + c\right)^{3}
$$
by \texttt{<function zero\_side>}, it remains to prove
$$
0 \leq -a^{3} - 24abc - b^{3} - c^{3} + \left(a + b + c\right)^{3}
$$
by \texttt{<function try\_expand\_r>}, it remains to prove
$$
0 \leq 3a^{2}b + 3a^{2}c + 3ab^{2} - 18abc + 3ac^{2} + 3b^{2}c + 3bc^{2}
$$
by \texttt{<function check\_AM\_GM>}, it remains to prove
$$
0 \leq 3a^{2}b - 9abc + 3ac^{2} + 3b^{2}c
$$
by \texttt{<function check\_AM\_GM>}, this is true!
\end{small}
\end{tcolorbox}
%
\end{figure}
\vspace{36pt}

\vspace{24pt}

\begin{figure}[!h]
\vspace{-0.3in}
\begin{tcolorbox}[colback=orange!8!]
\begin{small}
\textbf{6. Solution to USAMO-2003-P5.}\\
To prove
$$\frac{\left(a + b + 2 c\right)^{2}}{2 c^{2} + \left(a + b\right)^{2}} + \frac{\left(a + 2 b + c\right)^{2}}{2 b^{2} + \left(a + c\right)^{2}} + \frac{\left(2 a + b + c\right)^{2}}{2 a^{2} + \left(b + c\right)^{2}} \leq 8$$
we have
$$f(x) = \dfrac{(x+1)^2}{(1-x)^2+2x^2} \leq \dfrac{12x+4}{3} \text{ for } 0<x<1$$

$$\iff- \dfrac{\left(3 x - 1\right)^{2} \cdot \left(4 x + 1\right)}{3 \cdot \left(3 x^{2} - 2 x + 1\right)} \leq 0 \text{ for } 0<x<1,$$ which is true.\\
Substitute $x$ for $\dfrac{c}{a+b+c}$, we have
$$\frac{\left(a + b + 2 c\right)^{2}}{2 c^{2} + \left(a + b\right)^{2}}\leq \frac{4 c}{a + b + c} + \frac{4}{3}$$
It remains to prove
$$\frac{4 a}{a + b + c} + \frac{4 b}{a + b + c} + \frac{4 c}{a + b + c} + 4 \leq 8$$
by \texttt{<function try\_together\_l>}, this is true!
\end{small}
\end{tcolorbox}
%
\end{figure}

\begin{figure}[!h]
\vspace{-0.3in}
\begin{tcolorbox}[colback=orange!8!]
\begin{small}
\textbf{7. Solution to Polish-2004 Problem}\\
We use H\"older's inequality:
\[(a+b+c+d)^4 \leq (\frac{a}{(a^3 + 63bcd)^{\frac{1}{3}}} + \frac{b}{(63acd + b^3)^{\frac{1}{3}}} + \frac{c}{(63abd + c^3)^{\frac{1}{3}}} + \frac{d}{(63abc + d^3)^{\frac{1}{3}}})^3\times\]\[(a(a^3+63bcd)+b(b^3+63acd)+c(c^3+63abd)+d(d^3+63abc)).\]
It remains to prove $$1\leq \dfrac{(a+b+c+d)^\frac{4}{3}}{(a^4+252abcd+b^4+c^4+d^4)^\frac{1}{3}},$$
by \texttt{<function no\_pow>}, it remains to prove $$1\leq \dfrac{(a+b+c+d)^4}{a^4+252abcd+b^4+c^4+d^4},$$
by \texttt{<function nodiv\_expr>}, it remains to prove $$a^4+252abcd+b^4+c^4+d^4\leq (a+b+c+d)^4,$$
by \texttt{<function zero\_side>}, it remains to prove
$$0 \leq - a^{4} - 252 a b c d - b^{4} - c^{4} - d^{4} + \left(a + b + c + d\right)^{4}$$
by \texttt{<function try\_expand\_r>}, it remains to prove
\[0 \leq 4 a^{3} b + 4 a^{3} c + 4 a^{3} d + 6 a^{2} b^{2} + 12 a^{2} b c + 12 a^{2} b d + 6 a^{2} c^{2} + 12 a^{2} c d + 6 a^{2} d^{2} + 4 a b^{3}\dots\]
by \texttt{<function check\_AM\_GM>}, it remains to prove
\[0 \leq 4 a^{3} b + 4 a^{3} c + 4 a^{3} d + 6 a^{2} b^{2} + 12 a^{2} b c + 12 a^{2} b d + 12 a^{2} c d + 6 a^{2} d^{2} + 4 a b^{3}\dots\]

by \texttt{<function sep\_neg>}, it remains to prove
\[216 a b c d \leq 4 a^{3} b + 4 a^{3} c + 4 a^{3} d + 6 a^{2} b^{2} + 12 a^{2} b c + 12 a^{2} b d + 12 a^{2} c d + 6 a^{2} d^{2}\dots\]
by \texttt{<function check\_AM\_GM>}, it remains to prove
\[216 a b c d \leq 4 a^{3} b + 4 a^{3} c + 4 a^{3} d + 12 a^{2} b c + 12 a^{2} b d + 12 a^{2} c d + 4 a b^{3} + \dots\]
by \texttt{<function check\_AM\_GM>}, it remains to prove
\[216 a b c d \leq 4 a^{3} b + 4 a^{3} c + 12 a^{2} b c + 12 a^{2} b d + 12 a^{2} c d + 12 a b^{2} c + 12 a b^{2} d + 12 a b c^{2} +\dots\]
by \texttt{<function check\_AM\_GM>}, it remains to prove
\[216 a b c d \leq 4 a^{3} b + 4 a^{3} c + 12 a^{2} b c + 12 a^{2} c d + 12 a b^{2} d + 12 a b c^{2} + 88 a b c d + 12 a b d^{2} +\dots\]
by \texttt{<function check\_AM\_GM>}, it remains to prove
\[216 a b c d \leq 4 a^{3} b + 4 a^{3} c + 12 a^{2} b c + 136 a b c d + 12 a b d^{2} + 4 a c^{3} + 12 a c^{2} d\]
\[+ 4 a d^{3} + 4 b^{3} c + 4 b^{3} d + 12 b^{2} c d + 4 b d^{3} + 4 c^{3} d\]
by \texttt{<function check\_AM\_GM>}, it remains to prove
$$216 a b c d \leq 4 a^{3} b + 4 a^{3} c + 184 a b c d + 4 a c^{3} + 4 a d^{3} + 4 b^{3} c + 4 b^{3} d + 4 b d^{3} + 4 c^{3} d$$
by \texttt{<function zero\_side>}, it remains to prove
$$0 \leq 4 a^{3} b + 4 a^{3} c - 32 a b c d + 4 a c^{3} + 4 a d^{3} + 4 b^{3} c + 4 b^{3} d + 4 b d^{3} + 4 c^{3} d$$
by \texttt{<function check\_AM\_GM>}, it remains to prove
$$0 \leq 4 a^{3} b - 16 a b c d + 4 a d^{3} + 4 b^{3} c + 4 c^{3} d$$
by \texttt{<function check\_AM\_GM>}, this is true!
\end{small}
\end{tcolorbox}
%
\end{figure}


\begin{figure}[!h]
\vspace{-0.3in}
\begin{tcolorbox}[colback=orange!8!]
\begin{small}
\textbf{8. Solution to USA-IMO-Team-Selection-2010-P2.}\\
By \texttt{<function try\_homo>}, it is equivalent to prove
$$\frac{a^{3} b^{3}}{c^{2} \left(a + 2 b\right)^{2}} + \frac{a^{3} c^{3}}{b^{2} \left(2 a + c\right)^{2}} + \frac{b^{3} c^{3}}{a^{2} \left(b + 2 c\right)^{2}} \geq \frac{a^{\frac{2}{3}} b^{\frac{2}{3}} c^{\frac{2}{3}}}{3}$$
we use H\"older's inequality: 
\[(ab + ac + bc)^3 \leq (a(b + 2c) + b(2a + c) + c(a + 2b))^2\times\]
\[(a^3b^3/(c^2(a + 2b)^2) + a^3c^3/(b^2(2a + c)^2) + b^3c^3/(a^2(b + 2c)^2)).\]
It remains to prove
$$\frac{a^{\frac{2}{3}} b^{\frac{2}{3}} c^{\frac{2}{3}}}{3} \leq \frac{a b}{9} + \frac{a c}{9} + \frac{b c}{9}$$
by \texttt{<function check\_AM\_GM>}, this is true!
\end{small}
\end{tcolorbox}
%
\end{figure}

\begin{figure}[!h]
\vspace{-0.3in}
\begin{tcolorbox}[colback=orange!8!]
\begin{small}
\textbf{9. Solution to Korea-2011-P4.}\\
To prove $$ \frac{1}{a^2 - 4a + 9} + \frac{1}{b^2 - 4b + 9} + \frac{1}{c^2 - 4c + 9} \leq \frac{7}{18},$$
we have $$f(x)=1/(x^2-4x+9)\leq \dfrac{2+x}{18} \text{ for } 0<x<1$$

$$\iff\,\,\,- \frac{x \left(x - 1\right)^{2}}{18 \left(x^{2} - 4 x + 9\right)}\leq 0 \text{ for } 0<x<1,$$ which is true.
Substitute $x$ for $a/(a+b+c)$, we have
$$1/(a^2-4a+9)=\frac{\left(a + b + c\right)^{2}}{a^{2} - 4 a \left(a + b + c\right) + 9 \left(a + b + c\right)^{2}}\leq \frac{3a+2b+2c}{18a+18b+18c}.$$
It remains to prove $$\frac{3a+2b+2c}{18a+18b+18c}+\frac{2a+3b+2c}{18a+18b+18c}+\frac{2a+2b+3c}{18a+18b+18c}\leq \frac{7}{18},$$
by \texttt{<function try\_together\_l>}, this is true.
\end{small}
\end{tcolorbox}
%
\end{figure}

\begin{figure}[!h]
\vspace{-0.3in}
\begin{tcolorbox}[colback=orange!8!]
\begin{small}
\textbf{10. Solution to Japan-2014-P5}\\
By \texttt{<function try\_homo>}, it is equivalent to prove \[\frac{a \left(a + b + c\right)}{9 b c + 4 \left(b - c\right)^{2} + \left(a + b + c\right)^{2}} + \frac{b \left(a + b + c\right)}{9 a c + 4 \left(- a + c\right)^{2} + \left(a + b + c\right)^{2}} +\]
\[\frac{c \left(a + b + c\right)}{9 a b + 4 \left(a - b\right)^{2} + \left(a + b + c\right)^{2}} \geq \frac{1}{2}.\]
We use H\"older's inequality:
\begin{scriptsize}
    
\[\left(a + b + c\right)^{3}\leq \]
\[(\frac{a \left(a + b + c\right)}{9 b c + 4 \left(b - c\right)^{2} + \left(a + b + c\right)^{2}} + \frac{b \left(a + b + c\right)}{9 a c + 4 \left(- a + c\right)^{2} + \left(a + b + c\right)^{2}} + \frac{c \left(a + b + c\right)}{9 a b + 4 \left(a - b\right)^{2} + \left(a + b + c\right)^{2}})\times\]
\[\Bigl\{ a\left(9 a^{2} + 4 \left(b - c\right)^{2} + \left(a + b + c\right)^{2}\right) + b \left(9 b^{2} + 4 \left(- a + c\right)^{2} + \left(a + b + c\right)^{2}\right) +\]
\[c \left(9 c^{2} + 4 \left(a - b\right)^{2} + \left(a + b + c\right)^{2}\right)\Bigr\}\]
\end{scriptsize}

It remains to prove 
\begin{scriptsize}
$$\frac{1}{2} \leq \frac{\left(a + b + c\right)^{3}}{27 a b c + 4 a \left(b - c\right)^{2} + a \left(a + b + c\right)^{2} + 4 b \left(a - c\right)^{2} + b \left(a + b + c\right)^{2} + 4 c \left(a - b\right)^{2} + c \left(a + b + c\right)^{2}}$$
\end{scriptsize}
by \texttt{<function nodiv\_expr>}, it remains to prove
\[27 a b c + 4 a \left(b - c\right)^{2} + a \left(a + b + c\right)^{2} + 4 b \left(a - c\right)^{2} + b \left(a + b + c\right)^{2} + 4 c \left(a - b\right)^{2} + c \left(a + b + c\right)^{2}\]
\[\leq 2 \left(a + b + c\right)^{3}\]
by \texttt{<function zero\_side>}, it remains to prove
\[0 \leq - 27 a b c - 4 a \left(b - c\right)^{2} - a \left(a + b + c\right)^{2} - 4 b \left(a - c\right)^{2} - b \left(a + b + c\right)^{2} - 4 c \left(a - b\right)^{2}\]
\[- c \left(a + b + c\right)^{2} + 2 \left(a + b + c\right)^{3}\]
by \texttt{<function try\_expand\_r>}, it remains to prove
$$0 \leq a^{3} - a^{2} b - a^{2} c - a b^{2} + 3 a b c - a c^{2} + b^{3} - b^{2} c - b c^{2} + c^{3}$$
by \texttt{<function check\_schur>}, this is true!
\end{small}
\end{tcolorbox}
%
\end{figure}

\clearpage
\newpage
\section{Human Evaluation of Generated Synthetic Theorems}
\label{syn-thms}
We select 10 synthetic problems generated by our AIPS for evaluation, and 4 IMO problems for comparison. Then, we invite three professional contestants to evaluate the difficulty and elegance of these 14 problems. Two of the evaluators are National Mathematical Olympiad gold medalists, and one is a silver medalist.
The difficulty and elegance are needed to assign a score from 1 to 7, respectively.

\subsection{10 Synthetic Theorems and 4 Comparison IMO Problems}
\subsubsection{10 Synthetic Theorems}
\begin{itemize}
    \item \textbf{(Problem1)}\\
    Given $a,b,c>0$, then \\
    $\dfrac{(a+b+c)^3}{(ab+bc+ca)^2}\leq \dfrac{4a}{(b+c)^2}+\dfrac{4b}{(c+a)^2}+\dfrac{4c}{(a+b)^2}$
    \item \textbf{(Problem2)}\\
    Given $a,b,c>0$, then \\
    $\dfrac{27(a^2+b^2)^2(b^2+c^2)^2(c^2+a^2)^2}{(a^4+b^4+c^4+3a^2b^2+3b^2c^2+3c^2a^2)^3}\leq 1$
    \item \textbf{(Problem3)}\\
    Given $a,b,c>0$, then \\
    $\dfrac{abc(a+b+c)^3}{3(ab+bc+ca)(a^3c+ab^3+bc^3)}\leq 1$
    \item \textbf{(Problem4)}\\
    Given $a,b,c>0$, then \\
    {\footnotesize$\dfrac{2a}{\sqrt{2a^2+b^2+c^2}}+\dfrac{2b}{\sqrt{2b^2+c^2+a^2}}+\dfrac{2c}{\sqrt{2c^2+a^2+b^2}}\leq \dfrac{3\sqrt{2}(a+b+c)}{\sqrt{5a^2+5b^2+5c^2+ab+bc+ca}}$}
    \item \textbf{(Problem5)}\\
    Given $a,b,c>0$, then \\
    {$\dfrac{\sqrt{6}(a+b+c)^2}{6\sqrt{a^4+b^4+c^4+a^2b^2+b^2c^2+c^2a^2}}\leq \dfrac{a}{\sqrt{2a^2+b^2+c^2}}+\dfrac{b}{\sqrt{2b^2+c^2+a^2}}+\dfrac{c}{\sqrt{2c^2+a^2+b^2}}$}
    \item \textbf{(Problem6)}\\
    Given $a,b,c>0$, then \\
    $2(a+b+c)^{\frac{3}{2}} \leq (\sqrt{a+b} + \sqrt{b+c} + \sqrt{c+a}) \sqrt{a^2+b^2+c^2+ab+bc+ca}$
    \item \textbf{(Problem7)}\\
    Given $a,b,c>0$, then \\
    $\dfrac{(a^4+b^4+c^4)^\frac{3}{2}}{\sqrt{ab^2+bc^2+ca^2-abc} \sqrt{a+b+c}} \leq \dfrac{a^5}{\sqrt{ca+b^2}} + \dfrac{b^5}{\sqrt{ab+c^2}} + \dfrac{c^5}{\sqrt{bc+a^2}}$
    \item \textbf{(Problem8)}\\
    Given $a,b,c>0$, then \\
    $\dfrac{54 a b c + \left(a + b + c\right)^{3}}{\left (\sqrt{a^{2} + 2 b c} + \sqrt{2 a b + c^{2}} + \sqrt{2 a c + b^{2}}\right)^{2}} \leq a + b + c$
    \item \textbf{(Problem9)}\\
    Given $a,b,c>0$, then \\
    $\dfrac{a^{2} b}{\left(a + b\right)^{3}} + \dfrac{a c^{2}}{\left(a + c\right)^{3}} + \dfrac{b^{2} c}{\left(b + c\right)^{3}} \leq \dfrac{3}{8}$
    \item \textbf{(Problem10)}\\
    Given $a,b,c>0$, then \\
    $\dfrac{\left(a b + a c + b c\right)^{2}}{\sqrt{a^{2} + b^{2} + c^{2}} \sqrt{a^{2} + b^{2} + c^{2} + 3 a b + 3 b c + 3 c a}} \leq \dfrac{a^{2} b}{\sqrt{ b^{2} + 3 a c}} + \dfrac{b^{2} c}{\sqrt{ c^{2} + 3 a b}} + \dfrac{c^{2} a}{\sqrt{a^{2} + 3 b c}}$
\end{itemize}

\subsubsection{4 IMO Problems}

\begin{itemize}
    \item \textbf{(1995-imo-2)}\\ Given $a,b,c>0$ and $abc = 1$, then $\dfrac{1}{c^{3} \left(a + b\right)} + \dfrac{1}{b^{3} \left(a + c\right)} + \dfrac{1}{a^{3} \left(b + c\right)} \geq \dfrac{3}{2}$
    \item \textbf{(2001-imo-2)}\\ Given $a,b,c>0$, then $\dfrac{a}{\sqrt{a^{2} + 8 b c}} + \dfrac{b}{\sqrt{8 a c + b^{2}}} + \dfrac{c}{\sqrt{8 a b + c^{2}}} \geq 1$
    \item \textbf{(2006-imo-3)}\\ Assume $a,b,c \text{ are three real numbers, then}$ $|ab(a^2-b^2) + bc(b^2-c^2) + ca(c^2-a^2)| \leq \dfrac{9}{16\sqrt{2}}(a^2+b^2+c^2)^2$
    \item \textbf{(2020-imo-2)}\\ Assume $a\geq b\geq c\geq d\geq 0$ and $a+b+c+d=1$, prove that $a^{a} b^{b} c^{c} d^{d} \left(a + 2 b + 3 c + 4 d\right) < 1$
\end{itemize}

\subsection{Human Evaluation Results}

The rating scores by the three professional contestants are reported in Table \ref{tab:expert_scores}. The third expert does not assign scores to the four IMO problems, believing the average difficulty of the ten problems is significantly lower than that of IMO problems. The first expert does not give a difficulty score for Problem 8 because he does not solve it. From the table, we observe that while the average difficulty does not compare with IMO inequalities, a few problems, such as Problem 9 and Problem 7, reach the IMO level.

\begin{table}[h]
\centering
\caption{Scores given by human experts on synthetic theorems and IMO problems. Scores range from 1 to 7. \textbf{GM} denotes gold medalist, and \textbf{SM} denotes silver medalist.
}
\begin{tabular}{|c|c|c|c|c|c|c|}
\hline
\textbf{Problem} & \multicolumn{2}{c|}{\textbf{Expert 1 (GM)}} & \multicolumn{2}{c|}{\textbf{Expert 2 (GM)}} & \multicolumn{2}{c|}{\textbf{Expert 3 (SM)}} \\ \hline
 & \textbf{Difficulty} & \textbf{Elegance} & \textbf{Difficulty} & \textbf{Elegance} & \textbf{Difficulty} & \textbf{Elegance} \\ \hline
1 & 2 & 2 & 2 & 3 & 1 & 2.5 \\ \hline
2 & 1 & 1 & 1 & 2 & 1 & 1 \\ \hline
3 & 2 & 1 & 4 & 2 & 1.5 & 1 \\ \hline
4 & 3 & 2 & 3 & 2 & 2 & 1.5 \\ \hline
5 & 2 & 1 & 2 & 2 & 1.5 & 1 \\ \hline
6 & 2 & 2 & 2 & 2 & 1.5 & 1.5 \\ \hline
7 & 5 & 1 & 4 & 2 & 2 & 2 \\ \hline
8 & NA & 2 & 3 & 2 & 1 & 1.5 \\ \hline
9 & 4 & 3 & 4 & 5 & 2.5 & 2 \\ \hline
10 & 4 & 1 & 3 & 1 & 1 & 1.5 \\ \hline
IMO-1995-2 & 2 & 4 & 3 & 5 & NA & NA \\ \hline
IMO-2001-2 & 3 & 4 & 3 & 5 & NA & NA \\ \hline
IMO-2006-3 & 3 & 3 & 5 & 3 & NA & NA \\ \hline
IMO-2020-2 & 2 & 2 & 4 & 3 & NA & NA \\ \hline
\end{tabular}
\label{tab:expert_scores}
\end{table}

\newpage

\subsection{Synthetic Theorem Selected for Mathematical Olympiad}

\begin{figure}[h]
    \centering
    \begin{minipage}{0.45\textwidth}
        \centering
        \includegraphics[width=\textwidth]{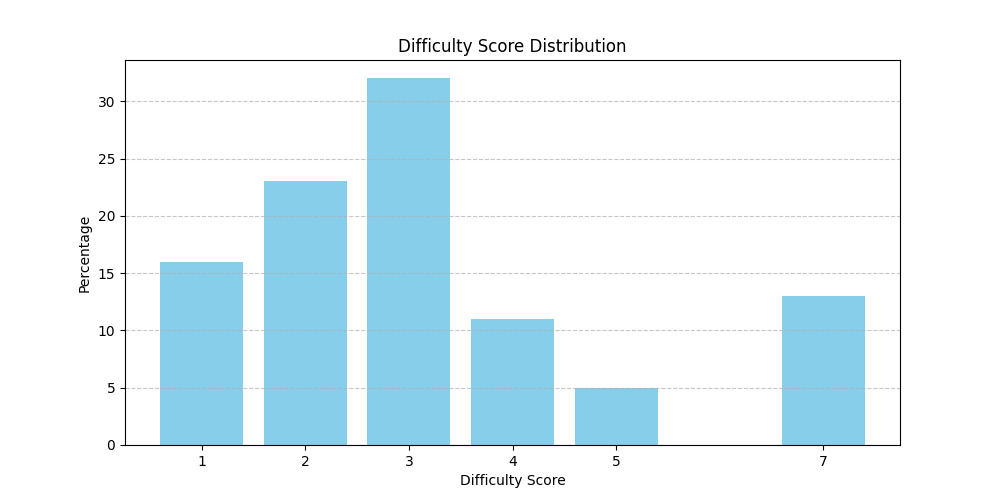}
    \end{minipage}
    \hfill
    \begin{minipage}{0.45\textwidth}
        \centering
        \includegraphics[width=\textwidth]{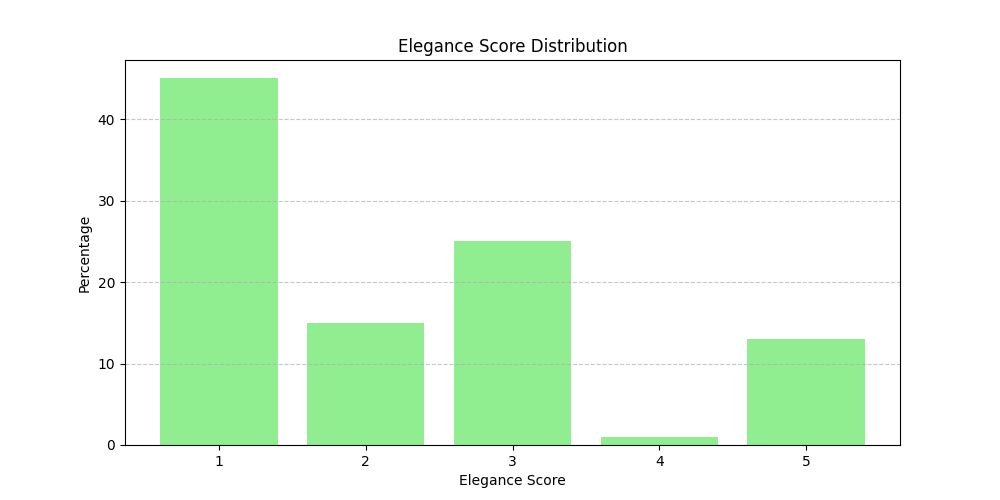}
    \end{minipage}
    \caption{Score distributions evaluated by 75 contestants online.}
    \label{fig:score_distribution}
\end{figure}

Among the 10 synthetic problems above, problem 4 was chosen as a competition problem in a major city's 2024 Mathematical Olympiad, as shown in Fig. \ref{fig:problem_statement}. It received positive feedback for its appropriate difficulty, concise form, and variety of solutions. This problem was posted online, and 75 contestants provided their evaluations on its difficulty and elegance. The score distributions are shown in Fig. \ref{fig:score_distribution}. The average difficulty score was 3.3 out of 7, and the elegance score was 2.2 out of 5. The 4 solutions to this problem, including one provided by our AIPS and 3 solutions collected from the competition organizers, are given as follows.

\begin{figure}[h]
\begin{tcolorbox}[colback=blue!8!,width=5in]
    \begin{minipage}{\linewidth}
        \centering
        \textbf{Problem:} Given three positive real numbers $a, b, c$, prove that 
        {\scriptsize 
        \[\dfrac{2a}{\sqrt{2a^2+b^2+c^2}}+\dfrac{2b}{\sqrt{2b^2+c^2+a^2}}+\dfrac{2c}{\sqrt{2c^2+a^2+b^2}}\leq \dfrac{3\sqrt{2}(a+b+c)}{\sqrt{5a^2+5b^2+5c^2+ab+bc+ca}}\]}
    \end{minipage}
\end{tcolorbox}
\caption{Selected theorem for a major city's Mathematical Olympiad.}
\label{fig:problem_statement}
\end{figure}

\begin{figure}[t]
\vspace{-0.3in}
\begin{tcolorbox}[colback=orange!8!,width=5in]
\begin{small}

\textbf{Proof 1. (Modified from AIPS' proof)} \\
$f''(x) = \dfrac{6x(-a^2-b^2-c^2)}{(a^2+b^2+c^2+x^2)^\frac{5}{2}} < 0$ for $x$ satisfying $0 < x < a^2+b^2+c^2$, where $f(x) = \dfrac{2x}{\sqrt{x^2+a^2+b^2+c^2}}$. By Jensen's inequality, \textbf{LHS} $\leq 3 \cdot \dfrac{2 \cdot \frac{a+b+c}{3}}{\sqrt{a^2+b^2+c^2+\left(\frac{a+b+c}{3}\right)^2}}$. It suffices to prove

$$3 \cdot \dfrac{2 \cdot \frac{a+b+c}{3}}{\sqrt{a^2+b^2+c^2+\left(\frac{a+b+c}{3}\right)^2}} \leq \dfrac{3\sqrt{2}(a+b+c)}{\sqrt{5a^2+5b^2+5c^2+ab+bc+ca}}.$$

Expanding the left-hand side, this is true. \qed
\end{small}
\end{tcolorbox}
\end{figure}

\begin{figure}[h]
\vspace{-0.3in}
\begin{tcolorbox}[colback=orange!8!,width=5in]
\begin{small}
\textbf{Proof 2. (Given by Humans)}\\
Without loss of generality, assume $a\geq b\geq c$ and $a^2+b^2+c^2=1$. Then the inequality in question is equivalent to 
\[\sum \dfrac{a}{\sqrt{1+a^2}} \leq \dfrac{3(a+b+c)}{\sqrt{9+(a+b+c)^2}}\]
Notice that \[\dfrac{a}{\sqrt{1+a^2}} = \sqrt{1-\dfrac{1}{1+a^2}} \geq \sqrt{1-\dfrac{1}{1+b^2}} = \dfrac{b}{\sqrt{1+b^2}}\]
By Chebyshev inequality, we get $$(\sum \sqrt{1+a^2})(\sum \dfrac{a}{\sqrt{1+a^2}}) \leq 3(a+b+c).$$
Then it suffices to prove 
    \[\sum \sqrt{1+a^2} \geq \sqrt{9+(a+b+c)^2}\]
which is equivalent to show $6+2\sum ab \leq 2\sum \sqrt{1+a^2}\sqrt{1+b^2}.$
Notice that 
\[1+ab\leq \sqrt{1+a^2}\sqrt{1+b^2} \iff 2ab\leq a^2+b^2\]
and the right-hand-side holds by AM-GM inequality. Therefore we have finished the proof.\qed
\end{small}
\end{tcolorbox}
\end{figure}

\begin{figure}[t]
\vspace{-0.3in}
\begin{tcolorbox}[colback=orange!8!,width=5in]
\begin{small}
\textbf{Proof 3. (Given by Humans)}\\
First we divide the proof into two subgoals:
\begin{equation}\label{leq_1}
    \frac{3\sqrt{2}(a+b+c)}{\sqrt{5a^2+5b^2+5c^2+ab+bc+ca}} \geq \dfrac{2(a+b+c)}{\sqrt{\dfrac{4}{3}(a^2+b^2+c^2)}}
\end{equation}
and
\begin{equation}\label{leq_2}
    \sum \frac{2a}{\sqrt{\frac{4}{3}(a^2+b^2+c^2)}} \geq \sum \frac{2a}{\sqrt{2a^2+b^2+c^2}}
\end{equation}
Where $\sum$ denotes cyclic summation.
The proof of (\ref{leq_1}) follows from the fact that $a^2+b^2+c^2 \geq ab+bc+ca$. For the second part, we apply Chebyshev's inequality.

Without loss of generality, we assume $a\geq b\geq c$. First notice that

\begin{equation}\label{leq_3}
    \sum \frac{2a}{\sqrt{\frac{4}{3}(a^2+b^2+c^2)}} - \frac{2a}{\sqrt{2a^2+b^2+c^2}} = \frac{1}{3}\sum x_a(2a^2-b^2-c^2)
\end{equation}

where 
$$x_a = \frac{2a}{\sqrt{\frac{4}{3}(a^2+b^2+c^2)}\sqrt{2a^2+b^2+c^2}(\sqrt{\frac{4}{3}(a^2+b^2+c^2)}+\sqrt{2a^2+b^2+c^2)}}$$

 and $x_b$, $x_c$ are defined similarly. We claim that $x_a \geq x_b\geq x_c$. For $x_a \geq x_b$, it suffice to show two inequalities: 
 \[a\sqrt{a^2+2b^2+c^2}\geq b\sqrt{2a^2+b^2+c^2}\] 
 \[a(a^2+2b^2+c^2)\geq b(2a^2+b^2+c^2)\] 
 Both can be proven by factorization, and the proof of  $x_b \geq x_c$ is similar.

Since $a\geq b\geq c$, we get $2a^2-b^2-c^2\geq 2b^2-c^2-a^2\geq 2c^2-a^2-b^2$. Combining with $x_a\geq x_b\geq x_c$ and applying Chebyshev's inequality, we get $\sum x_a(2a^2-b^2-c^2) \geq 0$. Finally, combining with (\ref{leq_3}), we conclude that (\ref{leq_2}) is proved.\qed
\end{small}
\end{tcolorbox}
\end{figure}

\begin{figure}[h]
\vspace{-0.3in}
\begin{tcolorbox}[colback=orange!8!,width=5in]
\begin{small}
\textbf{Proof 4. (Given by Humans)}\\
    Let $S=a^2+b^2+c^2$ and $t = \dfrac{a+b+c}{3}$. Substituting into the inequality and rearranging:
    \[\mathrm{LHS} = \sum \dfrac{2a}{\sqrt{S+a^2}}\] 
    \[\mathrm{RHS} = \sum ((2(t^2+S)^{-\frac{1}{2}}-2t^2(t^2+S)^{-\frac{3}{2}})(a-t)+2t(t^2+S)^{-\frac{1}{2}})\]
    It suffice to show $$\dfrac{2a}{\sqrt{a^2+S}}\leq \dfrac{2S(a-t)+2t(t^2+S)}{(t^2+S)^\frac{3}{2}}$$
    which is equivalent to 
    \[3a^2t^4S+3a^2t^2S^2\leq (2Sa^3t^3+St^6)+(2S^2ta^3+S^2a^4)\]
    The last inequality is proved by applying AM-GM inequality.\qed
\end{small}
\end{tcolorbox}
\end{figure}

\clearpage




\end{document}